\documentclass[journal]{IEEEtran}

\usepackage{graphicx}
\usepackage{comment}
\usepackage{amsmath,amssymb} % define this before the line numbering.
\usepackage{color}
\usepackage{amsmath}
\usepackage{footnote}
\usepackage{xcolor,soul,framed} 
\usepackage{xspace}
\usepackage{array}
\usepackage{eqparbox}
\usepackage{url}
\usepackage{longtable}
\usepackage{lipsum}
\usepackage{blindtext}
\usepackage{makecell}
\usepackage{mathtools}
\usepackage{commath}
\usepackage{multirow}
\usepackage{tabularx}
\usepackage[normalem]{ulem}
\usepackage{xspace}
\usepackage{algorithm}
\usepackage{algpseudocode}
\usepackage{amsfonts}
\usepackage{subcaption}
\usepackage[symbol]{footmisc}

\usepackage{amssymb}% http://ctan.org/pkg/amssymb
\usepackage{pifont}% http://ctan.org/pkg/pifont
\newcommand{\cmark}{\ding{51}}%
\newcommand{\xmark}{\ding{55}}%
\newcommand{\etal}{\emph{et al. }}

\begin{document}

\title{Depth as Attention for Face Representation Learning}

\author{Hardik Uppal, \IEEEmembership{Student Member, IEEE}, Alireza Sepas-Moghaddam \IEEEmembership{Member, IEEE}, Michael Greenspan \IEEEmembership{Member, IEEE}, and Ali Etemad, \IEEEmembership{Senior Member, IEEE}% <-this % stops a space
\thanks{This work was funded by Irdeto Canada Corporation and the Natural Sciences and Engineering Research Council of Canada (NSERC). 

\noindent H. Uppal, A. Sepas-Moghaddam, M. Greenspan, and A. Etemad are with the Department of Electrical and Computer Engineering, Queen's University, Kingston,
ON, K7L 3N6 Canada (e-mail: hardik.uppal@queensu.ca).}}

\markboth{IEEE Transactions on Information Forensics and Security}%
{Shell \MakeLowercase{\textit{et al.}}: Bare Demo of IEEEtran.cls for IEEE Journals}

\maketitle
\begin{abstract}
Face representation learning solutions have recently achieved great success for various applications such as verification and identification. However, face recognition approaches that are based purely on RGB images rely solely on intensity information, and therefore are more sensitive to facial variations, notably pose, occlusions, and environmental changes such as illumination and background. A novel depth-guided attention mechanism is proposed for deep multi-modal face recognition using low-cost RGB-D sensors. Our novel attention mechanism directs the deep network ``where to look'' for visual features in the RGB image by focusing the attention of the network using depth features extracted by a Convolution Neural Network (CNN). The depth features help the network focus on regions of the face in the RGB image that contain more prominent person-specific information. Our attention mechanism then uses this correlation to generate an attention map for RGB images from the depth features extracted by the CNN. We test our network on four public datasets, showing that the features obtained by our proposed solution yield better results on the Lock3DFace, CurtinFaces, IIIT-D RGB-D, and KaspAROV datasets which include challenging variations in pose, occlusion, illumination, expression, and time lapse. Our solution achieves average (increased) accuracies of 87.3\% (+5.0\%), 99.1\% (+0.9\%), 99.7\% (+0.6\%) and 95.3\%(+0.5\%) for the four datasets respectively, thereby improving the state-of-the-art. \textcolor{black}{We also perform additional experiments with thermal images, instead of depth images, showing the high generalization ability of our solution when adopting other modalities for guiding the attention mechanism instead of depth information}.
\end{abstract}


\section{Introduction} \label{sec:intro}
% ---> FR general introduction
Face recognition (FR) systems have been successfully used for human identification with very high accuracy and generalizability \cite{jain}. Since the emergence of the first FR system around half a century ago~\cite{goldstein1971identification}, this area has witnessed significant progress, notably benefiting more recently from the advances in deep neural networks (DNNs) \cite{masi2018deep}. Nowadays, DNNs such as convolutional neural networks (CNNs) have opened a new range of possibilities for designing improved FR methods, and have dominated the state-of-the-art for both verification (one-to-one) and identification (one-to-many) tasks~\cite{CNNFR}. Despite these recent advances, certain common conditions such as changes in lighting, viewing angles, and non-uniform backgrounds, as well as changes in human appearance due to aging, emotions, and occlusions, still limit FR performance~\cite{faceS}. 
% To address these challenges, DNNs must be trained using millions of face images, which encompass these variations~\cite{masi2018deep}.    

% ---> FR using RGB-D
The emergence of new types of imaging sensors has also opened new frontiers for FR systems~\cite{mobile}. For example, multi-modal RGB-D (red, green, blue, and depth) cameras, such as the consumer-level Microsoft Kinect~\cite{zhang2012microsoft} and Intel RealSense~\cite{Keselman_2017_CVPR_Workshops} sensors, have made it possible and cost-effective to simultaneously capture co-registered color intensity and depth data of a scene~\cite{RGBD}. Depth (range) information can be instrumental for FR, as it provides geometric information about the face, in the form of dense 3D points that sample the surface of facial components~\cite{RGBDFR}.

\begin{figure}[!t]          
\centering
\includegraphics[width=0.8\columnwidth]{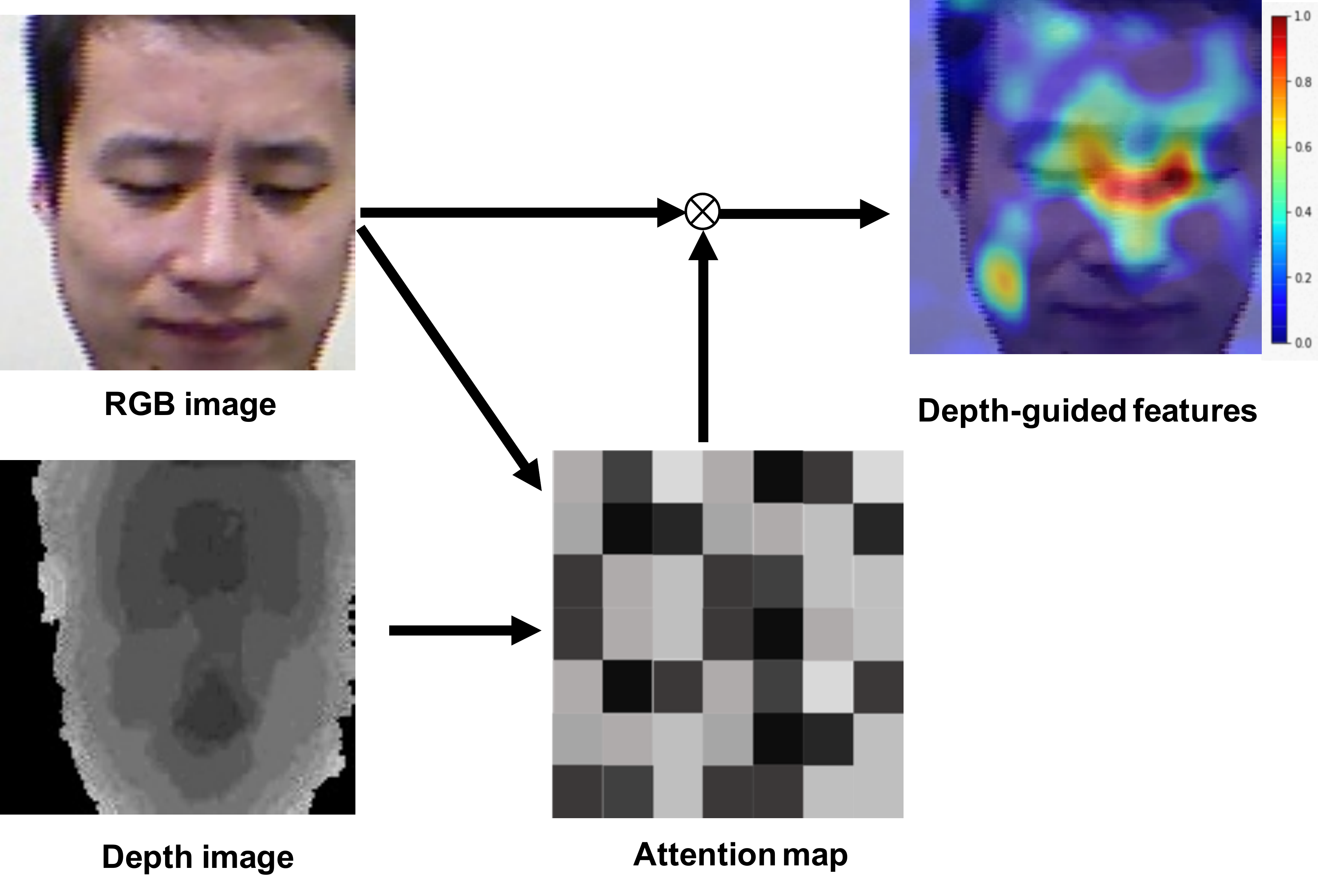}
\caption{An overview of our proposed depth-guided attention is presented. The depth and the RGB images are used to compute an attention map that is then used to focus on the most important person-specific parts available in RGB features for FR.}
\label{fig:attention}
\end{figure}

% ---> Limitations with current methods
Current models for recognition often process the input facial images uniformly~\cite{CBAM,chen2017sca}, applying similar attention toward different regions of the image. This is despite the fact that certain facial regions like the eyes, mouth, nose, cheeks, ear, and  chin~\cite{abudarham2019critical,ellis1979identification,kramer2018familiarity, eye, ear} are known to contain a high degree of person-specific information that greatly inform face representation learning, particularly for identity recognition. To address this, methods have been proposed to focus the attention of the model on specific regions of the face or the learned embedding, often referred to as attention mechanisms~\cite{attsurvey,softatten}. Such methods have been shown to enhance the performance of face representation learning models resulting in higher accuracies in applications such as FR~\cite{rao2017attention} and presentation attack detection~\cite{wang2019multi}.

% ----> Our proposed method introduced
Interestingly, we observe a considerable amount of depth variation in the regions of the face mentioned above that contain important person-specific information. Additionally, depth information is known to be less sensitive to facial variations, notably pose, occlusions, and environmental variations such as illumination and background  \cite{depthvariation}. To exploit this characteristic, in this paper, we propose a novel attention mechanism, called \textit{depth-guided attention}, to apply different amounts of focus on different parts of the face based on depth variations. 
% Our solution, called \textit{depth-guided attention}, is inspired by the notion that certain regions of the face such as the lips, chin, and cheeks are likely to contain greater amounts of person-specific information. 
In our network, we first extract a set of feature maps from both RGB and depth images using the VGG~\cite{28} convolutional feature extractor. We then use the output convolutional maps to learn the similarities between the two feature maps using our feature pooling module. Then, an attention refinement module creates attention maps to highlight features with person-specific depth-related information as illustrated in Figure~\ref{fig:attention}.  

The performance of our solution has been tested on four prominent RGB-D face datasets, including Lock3DFace~\cite{7}, CurtinFaces~\cite{mian}, IIIT-D RGB-D~\cite{1,2}, and KaspAROV~\cite{chowdhury2016rgb_newadd,chhokra2018unconstrained_newadd}, and has been compared to a number of state-of-the-art RGB-D FR methods~\cite{13,11,ICIP}. The results reveal that our proposed solution consistently learns better person-specific face representations as evidenced by the improved performance in different FR tasks under different challenging conditions.

% ----> Contributions
Our key contributions are summarized as follows:
\begin{itemize}
    \item \textcolor{black}{We propose a novel attention mechanism for face representation learning, exploiting co-registered RGB and depth images to selectively focus on important salient facial features in the input RGB image. This is achieved by finding the similarity between the feature embeddings of RGB and depth images, and computing an attention map for the RGB input. This module could be trained in an end-to-end trainable fashion with a convolutional network as a feature extractor, as long as an additional modality exists to guide the attention};
    % \item Our proposed method can directly process the input RGB-D images without the need to perform a (possibly costly) pre-processing step;    
    % \item The proposed approach is highly dynamic as it generates an attention map that is fine-tuned to the input RGB image, and does not require human supervision to determine the information-rich parts of the face;
    % \item Our proposed solution sets new \textbf{state-of-the-art} results on four public and large RGB-D FR datasets;
    \item \textcolor{black}{Our proposed solution shows superior recognition performance in dealing with challenging scenarios such as face pose, occlusions, illuminations, and expression when compared to other fusion strategies including feature or score-level fusions, as well as a number of other custom-designed solutions in the literature. Moreover, our method sets new \textbf{state-of-the-art} results on four public and large RGB-D FR datasets;}
    \item \textcolor{black}{Our proposed solution shows a high generalization ability when adopting thermal images, instead of the depth images, to guide the attention module.}
    % Our proposed solution shows robust performance in dealing with challenging scenarios such as facial and environmental variations.
\end{itemize}
% ----> paper organisation

The rest of the paper is organized as follows: Section~\ref{sec:Related work} provides an overview of RGB-D face datasets and recognition methods. Our FR solution based on the proposed depth-guided attention mechanism is presented in Section~\ref{sec:Method}. Section~\ref{sec:Experiment} presents the experimental setup, and Section~\ref{sec:Results} the results of  comparing the proposed method with other state-of-the-art RGB-D FR methods, along with an analysis of their performance and generalization ability. Section~\ref{sec:Conclusion} concludes the paper with a summary and a discussion of future work.

\section{Related work}\label{sec:related_work}
\label{sec:Related work}
% -----> RGB-D classical methods
\subsection{RGB-D Face Recognition Methods}
RGB-D FR methods can be classified into hand-crafted based and deep learning based categories~\cite{faceS}. Table~\ref{tab2} overviews the main characteristics of RGB-D FR methods, sorted chronologically according to their release dates. This table highlights the FR categories as well as the feature extractors and the classifiers used for recognition. This table also presents the strategies used to fuse RGB and depth information that can be carried out at several levels, of which the feature-level and score-level fusion strategies are the most often employed~\cite{Fusion}.

\begin{table*}
  \centering
%   \setlength\tabcolsep{2pt}
%   \scriptsize
    % \footnotesize
    \caption{Overview of state-of-the-art RGB-D FR methods.}
    \vspace{2mm}
    \begin{tabular}{l l l l l l l}
    \hline
    \textbf{Ref.}& \textbf{Year}& \textbf{Cat.}& \textbf{Feature Extractor} & \textbf{Classifier} & \textbf{Fusion} & \textbf{Dataset} \\
    \hline
     \cite{1}& 2013 & Hand-crafted & HOG & RDF & Feature-level & IIIT-D \\
     \cite{mian}& 2013 & Hand-crafted & ICP, DCS & SRC & N/A & IIIT-D \\
      \cite{5} & 2014 & Hand-crafted & PCA, LBP, SIFT, LGBP & kNN & Score-level & Kinect Face \\
     \cite{2} & 2014 & Hand-crafted & RISE+HOG & RDF & Feature-level & IIIT-D \\
     \cite{7} & 2016 & Hand-crafted & ICP & SDF & N/A & Lock3DFace \\
     \cite{29} & 2016 & Hand-crafted & Covariance matrix rep. & SVM & Score-level & CurtinFaces \\
     \cite{chowdhury2016rgb_newadd} & 2016 & Deep learning & Autoencoder & Softmax & Score-level & Kinect Face \\
     \cite{9} & 2018 & Deep learning & Siamese CNN & Softmax & Feature-level & Pandora \\
     \cite{13} & 2018 & Deep learning & 9 Layers CNN + Inception & Softmax & Feature-level & VAP, IIIT-D, Lock3DFace \\
     \cite{LFFace} & 2018 & Deep learning & Fine-tuned VGG-Face & Softmax & Feature-level & LFFD \\
     \textcolor{black}{\cite{Jiang_PAMI}} & \textcolor{black}{2018} & \textcolor{black}{Deep learning} & \textcolor{black}{Custom CNN} & \textcolor{black}{Attribute-aware loss} & \textcolor{black}{Feature-level} & \textcolor{black}{Private dataset} \\
     \textcolor{black}{\cite{cui2018improving_newadd}} & \textcolor{black}{2018} & \textcolor{black}{Deep learning} & \textcolor{black}{Inception-v2} & \textcolor{black}{Softmax} & \textcolor{black}{Feature-level} & \textcolor{black}{IIIT-D, Lock3DFace} \\
     \cite{CVPR} & 2019 & Deep learning & 14 layers CNN + Attention & Softmax & Feature-level & Lock3DFace \\
     \cite{ICIP} & 2020 & Deep learning & CNN + two-level attention& Softmax & Feature-level & IIIT-D, CurtinFaces \\
     \textcolor{black}{\cite{lin2020rgb_newadd}} & \textcolor{black}{2020} & \textcolor{black}{Deep learning} & \textcolor{black}{Custom CNN} & \textcolor{black}{Assoc., Discrim., and Softmax} & \textcolor{black}{Feature-level} & \textcolor{black}{IIIT-D} \\
    \hline
    \end{tabular}%
  \label{tab2}%
\end{table*}%

The first RGB-D FR methods relied on hand-crafted visual descriptors. In one of the first attempts~\cite{1}, entropy maps corresponding to the input RGB and depth images along with a visual saliency map corresponding to the RGB image were computed. The Histogram of Oriented Gradients (HOG) descriptor was then applied, thus extracting features from these maps. The extracted features were finally concatenated to be used as input to a Random Decision Forest (RDF) classifier for recognizing identity. In~\cite{mian} the Iterative Closest Point (ICP) algorithm exploiting depth information was used for RGB face alignment. Discriminant Color Space (DCS) was then applied to the aligned RGB image, thus finding a set of linear combinations for the color components to maximize separability of the classes. Finally, a Sparse Representation Classifier (SRC) was used to perform FR. The performance of different feature extractors including Principle Component Analysis (PCA), Local Binary patterns (LBP), Scale-Invariant Feature Transform (SIFT), and Local Gabor Binary Patterns (LGBP) were compared for RGB-D FR, where LBP descriptor obtained the best performance. The RGB-D FR method proposed in~\cite{2} computed a new descriptor based on saliency and entropy maps, called RISE descriptor. Extracted features from different maps were concatenated and HOG descriptors were then used to provide an RDF classifier with the texture features. In the RGB-D FR method proposed in~\cite{7}, a 3D face model was reconstructed from RGB-D data using ICP, and a Signed Distance Function (SDF) was used to match the face models. In~\cite{29}, a block based covariance matrix representation was used to model RGB and depth images in a Riemannian manifold. Support Vector Machine (SVM) classification scores obtained from RGB and depth matrices were finally fused to perform FR.

% -----> RGB-D deep learning methods
As can be seen in Table~\ref{tab2}, the focus of RGB-D FR has shifted to deep learning methods since 2016. Various strategies have been used to make the most of depth information provided by RGB-D sensors. In~\cite{chowdhury2016rgb_newadd}, an RGB-D FR method was proposed based on an autoencoder architecture to learn a mapping function between RGB and depth modalities, thus generating a richer feature representation. A new training strategy was proposed in the context of RGB-D FR~\cite{9}, exploiting depth information to improve the learning of a distance metric during the training of a CNN. In~\cite{13}, a new architecture was used to learn from RGB and depth modalities, introducing a shared layer between two networks corresponding to the two modalities, thus allowing interference between modalities at early layers. In~\cite{LFFace}, RGB, disparity and depth images were independently used as inputs to a VGG-16 architecture for fine-tuning the VGG-Face model. The obtained embeddings were finally fused to feed an SVM classifier for performing FR. 
\textcolor{black}{
Jiang \etal\cite{Jiang_PAMI} presented an attribute-aware loss function for
CNN-based FR which aims to regularize the distribution of the learned feature vectors with respect to some soft-biometric attributes such as gender, ethnicity, and age, thus boosting FR results. Cui \etal\cite{cui2018improving_newadd} estimated the depth from RGB modality using a multi-task approach including face identification along with depth estimation. They also performed RGB-D recognition experiments to study the effectiveness of the estimated depth for the recognition task using the Inception-V2~\cite{ioffe2015batch} fusion network on the Lock3DFace and IIIT-D RGB-D public datasets. Lin \etal\cite{lin2020rgb_newadd} proposed an RGB-D face identification method by introducing new loss functions, including associative and discriminative losses, which were then combined with the softmax loss for training, showing boosted recognition results on the IIIT-D RGB-D dataset.
}

%----- 
%Some context for pose variation
\textcolor{black}{
Most of the challenging conditions in RGB-D face datasets relate to extreme pose variations and occlusions.
% as can be seen in our evaluation Table~\ref{tab:Lock3dFace}
Some RGB-based FR methods have attempted to solve extreme pose variations~\cite{zhao2018towards_newadd,zhao2017dual_newadd}. Zhao \etal\cite{zhao2017dual_newadd} proposed a dual-agent adversarial architecture, combining prior knowledge from the data distribution with adversarial training and also pose and identity perception losses in order to recover the lost information inherent in projecting a 3D face onto the 2D image space. In~\cite{zhao2018towards_newadd}, Zhao \etal presented a novel face frontalization network to be trained along with a FR network, thus learning pose-invariant representations for FR. Other challenging problems like occluded faces have been addressed in~\cite{zhao2017robust_newadd,song2019occlusion_newadd}. Song~\cite{song2019occlusion_newadd} used a pairwise differential siamese network between occluded and non-occluded faces to capture the correspondence between them. This information is then used to create a mask for the features which were occluded, thereby excluding those features from further processing
during recognition. Zhao~\etal\cite{zhao2017robust_newadd} used an LSTM autoencoder to remove facial
occlusions. Other works~\cite{li2017generative_newadd, iizuka2017globally_newadd} generated non-occluded images from occluded images utilizing adverserial learning, achieving very realistic results.}
% -----> general attention methods review

\subsection{Attention Mechanisms}
Human perception relies on attention, as proven in various studies~\cite{rensink2000dynamic,itti1998model}, to selectively concentrate on multiple entities that are available in a scene. This issue is more evident when the human brain tries to recognize human identity through facial images~\cite{Perception, Brain}. Attention mechanisms modeled after human perception have changed the way to work with CNNs~\cite{attsurvey}, by selectively focusing on the most important parts of the inputs, thus increasing the effective discrimination of the output embeddings. Attention mechanisms have so far been successfully used in different areas of computer vision and natural language processing~\cite{attsurvey,chaudhari2019attentive}. 
Soft attention mechanisms~\cite{softatten} have mostly been employed to selectively focus on the most important features extracted from multiple inputs, such as spatial features that have been extracted from video frames over the temporal sequence. In this context, the weights associated to each feature can be learned using a feed-forward neural network, and are ultimately multiplied by their corresponding features in order to obtain the resulting attention-refined features. 

% ----->RGB-D attention methods
Attention mechanisms have recently gained attention to exploit complementary RGB and depth information in the context of deep learning-based RGB-D FR methods. Mu~\etal~\cite{CVPR} proposed adding an attention weight map to each feature map, computed from RGB and depth modalities, thus focusing on the most important pixels with respect to their locations during training. Uppal~\etal~\cite{ICIP} used both spatial and channel information from depth and RGB images and fused the information using a two-step attention mechanism. The attention modules assign weights to features, choosing between features from depth and RGB and hence utilize the information from both data modalities effectively.

As a  special type of attention, \emph{cross-modal attention} or \emph{co-attention} mechanisms have been proposed, notably for visual question answering (VQA) and image captioning applications~\cite{xu2016ask,nam2017dual}. These methods jointly exploit the symmetry between one input vector and one reference vector (for example between images and questions in VQA) to guide the attention over the input vector. Various works have explored different ways to exploit this symmetry between input and reference vectors~\cite{luong2015effective, lin2015bilinear, kim2016hadamard}. The mechanism proposed in this paper is also a co-attention mechanism, where depth information is used to focus the attention of the network on specific parts of the RGB images. 

Capsule network~\cite{sabour2017dynamic} has recently been proposed that can also act as an attention mechanism~\cite{zhang2018attention,patrick2019capsule}. It is capable of learning feature importance by assigning more weights to the more relevant features while ignoring the spurious dimensions. A capsule network contains two main blocks, i.e. the \emph{primary capsule} and the \emph{high-level capsule}. The first primary capsule block encodes spatial information using convolutional layers, after which the second high-level capsule block then learns deeper part-whole relationships between hierarchical sub-parts. This network contains a trainable weight matrix for encoding the part-whole spatial relationships that can be considered as an attention layer over the transformed capsules. In this paper, we will compare the performance of other attention mechanisms, including soft and capsule attention, with our solution in Section~\ref{sec:Results}.

% ----->RGB-D datasets
\subsection{RGB-D Face Datasets}
RGB-D FR is a relatively new topic, and so only a few RGB-D face datasets have so far been made available. These datasets are generally collected in indoor environments under controlled settings, and are commonly referred to as \textit{constrained FR datasets}. Table~\ref{tab1} provides an overview of the characteristics of existing RGB-D face datasets including the type of sensors used, statistics about the size of the datasets, and the variations considered (i.e. different times, views, illuminations, expressions, and occlusions). These datasets are sorted in the table chronologically by release date.

\begin{table*}[!h]
  \centering
%   \setlength\tabcolsep{3pt}
%   \scriptsize
    % \footnotesize
    \caption{Overview of the available RGB-D face datasets with different characteristics.}
    \vspace{2mm}
    \begin{tabular}{ l l l l | l l | l l l l l}
    \hline
    \multicolumn{4}{ c |}{\textbf{Dataset}} & \multicolumn{2}{ c| }{\textbf{Statistics}} & \multicolumn{5}{ c }{\textbf{Face variations}} \\
    \hline    
    % \textbf \textbf{Name}& \textit{Year}& \textit{Sensor} &  \textit{\vtop{\hbox{\strut \# of}\hbox{\strut sub.}}} & \textit{\vtop{\hbox{\strut \# of}\hbox{\strut img.}}} & \textit{Age} & \textit{View}& \textit{Illum}.& \textit{Emot}.& \textit{Occl}.\\
        \textbf \textbf{Name} & \textit{Year} & \textit{Sensor}& \textit{Acquisition condition}&  \textit{\# of sub.} & \textit{\# of samp.} & \textit{Time} & \textit{View}& \textit{Illu}. & \textit{Expr.} & \textit{Occl}.\\
    \hline
    % Texas 3D []& 2010& {\vtop{\hbox{\strut Stereo}\hbox{\strut Vision}}} & 105 & 1149 &\xmark &\xmark&\xmark &\cmark &\xmark    \\
    BU-3DFE \cite{BU} & 2006 & 3dMD Scan. & Constrained & 100 & 2500 img. & \xmark & \xmark & \xmark & \cmark & \xmark    \\
    % \hline
    Texas 3D \cite{TX} & 2010& Stereo Vis. & Constrained &105 & 1149 img. &\xmark &\cmark&\xmark &\cmark &\xmark    \\
    % \hline
    VAP \cite{4} & 2012& Kinect I & Constrained &31 & 1149 img. & \xmark & \xmark & \xmark & \cmark & \xmark    \\
    % \hline
    IIIT-D \cite{1,2} & 2013 & Kinect I & Constrained &106 & 4605 img. & \xmark & \cmark & \cmark & \cmark & \xmark    \\
    % \hline
    CurtinFaces \cite{mian} & 2013 & Kinect I & Constrained &52 & 5000 img. & \xmark & \cmark & \cmark & \cmark & \cmark    \\
    % \hline
    % BU-3DFE  []& 2013& Stereo Vis. & 41 & 1149 &\xmark &\cmark&\xmark &\cmark &\xmark    \\
    FaceWarehouse \cite{FaceWarehouse} & 2014 & Kinect I &Constrained & 150 & 3000 img. & \xmark & \xmark & \xmark & \cmark & \xmark    \\
    % \hline
    Kinect Face \cite{5} & 2014 & Kinect I &Constrained & 52 & 936 img. & \cmark & \cmark & \cmark & \cmark & \cmark    \\
    % \hline
    LFFD \cite{LFFD} & 2016 & Light Field &Constrained & 100 & 4000 img. & \cmark & \cmark & \cmark & \cmark & \cmark    \\
    % \hline
    Lock3DFace   \cite{7} & 2016 & Kinect II &Constrained & 509 & 5711 vid. & \cmark & \cmark & \cmark & \cmark & \cmark \\
    \textcolor{black}{KaspAROV} \cite{chowdhury2016rgb_newadd, chhokra2018unconstrained_newadd} & \textcolor{black}{2016} & \textcolor{black}{Kinect I and II} & \textcolor{black}{Unconstrained} &\textcolor{black}{108} & \textcolor{black}{432 vid.} & \cmark & \cmark & \cmark & \cmark & \cmark    \\
    % \hline    

    \hline
    \end{tabular}%
  \label{tab1}%
\end{table*}%

Among the constrained RGB-D datasets listed in Table~\ref{tab1}, Texas 3D~\cite{TX}, Eurecom Kinect face~\cite{5}, and VAP~\cite{4} are relatively small, while BU-3DFE~\cite{BU} and FaceWarehouse~\cite{FaceWarehouse} are specifically designed for facial emotion recognition, which prevents their usage in our experiments. Additionally, it has been proven that the rendered depth images from light field multi-view data are not as effective as Kinect data for FR~\cite{VS}, so we also excluded LFFD~\cite{LFFD} from our experiments. To this end, we have conducted our experiments on the remaining three constrained datasets, i.e. IIIT-D~\cite{1,2}, CurtinFaces~\cite{mian}, and Lock3DFace~\cite{7}, which are also the largest available RGB-D face datasets. 

\textcolor{black}{Generally, RGB-D datasets containing challenging testing conditions with extreme poses, illumination, and expressions are also collected in constrained lab environments. In contrast, the KaspAROV~\cite{chowdhury2016rgb_newadd, chhokra2018unconstrained_newadd} dataset collected images in a surveillance-type setting, and is less constrained compared to other datasets. During collection, each subject walks back and forth within the field-of-view of the Kinect. No limitations are imposed on expression, pose, or gesture. Hence, the database contains unconstrained pose, illumination, and expression variations along with variations in capture distances. In order to show the efficiency of our proposed solution when dealing with unconstrained RGB-D face data, we have also included the KaspAROV~\cite{chowdhury2016rgb_newadd, chhokra2018unconstrained_newadd} dataset in our experiments. It must be noted that KaspAROV dataset is relatively small in terms of number of images and number of subjects as compared to the well-known in-the-wild RGB databases like Labelled Faces in the wild (LFW)~\cite{huang2008labeled} and YouTube faces (YTF)~\cite{wolf2011face}.}

\section{Method} \label{sec:Method}
% \subsection{Problem Definition and Model Overview}\label{sec:problem_definition}
\subsection{Model Intuition and Overview}\label{sec:problem_definition}
Certain facial regions like the eyes, mouth, nose, cheeks, ear, and  chin~\cite{abudarham2019critical,ellis1979identification,kramer2018familiarity, eye, ear} are known to contain a higher degree of person-specific information compared to other parts of the face. This property can be exploited, notably by using attention mechanisms, to improve the performance of FR systems by learning better person-specific representations. On the other hand, it is interestingly observed that the amount of depth variation in the mentioned important regions are more prominent. Additionally, depth information is less sensitive to variations such as pose, occlusions caused by face coverings, and environmental variations including illumination and background \cite{depthvariation}. Accordingly, focusing on facial regions containing important person-specific information can be more effective when exploiting depth information. These are the intuitions behind our proposal, called \textit{depth-guided attention}, which applies different amounts of focus on various parts of the RGB image based on depth variations. 

% Given a face recognition dataset $\mathcal{D}$, where each image $I_{i}$ is captured from identity $x_i \in \{x_1,x_2,\dots, x_N\}$, the task of RGB-D face recognition for a probe image, $I_{probe}$, can be defined as:
% \begin{equation} \label{eq:task}
%     \hat{x} = arg \max\limits_{x_i} Pr(x_i|I_{probe}, \mathcal{D}),
% \end{equation}
% where $Pr(x_i|I_{probe}, \mathcal{D})$ is the probability that image $I_{probe}$ belongs to identity $x_i$. 

Our depth-guided attention solution can be modeled according to Equation \ref{eq:model_preview} to produces a set of attention-refined features, $f$:
\begin{equation}\label{eq:model_preview}
    f = F_{RGB}\times DepthAtt(F_{RGB}, F_{Depth}),
\end{equation}
where $DepthAtt$ is the depth-guided attention mechanism that focuses the network based on the learned depth features, $F_{Depth}$. In other words, depth information effectively directs the deep network ``where to look'' for visual features within the RGB feature map, $F_{RGB}$.

%-------> architecture overview/data-flow
The overall architecture of our proposed network is shown in Figure~\ref{fig:architecture}. The network consists of two VGG-16 convolutional feature extractors whose outputs are the RGB and depth convolutional maps. These two networks are combined to form the \textbf{Convolutional Feature Extractor} module. The depth-guided attention mechanism is composed of the next two modules. In this context, first, we combine the extracted feature maps to create a pooled feature map through the \textbf{Feature Pooling} module. These pooled features are then fed to the \textbf{Attention Refinement} module to generate the attention for features extracted from the RGB branch. Finally the attention-refined features are fed to the \textbf{Classifier} module for recognition of the input identity.

%-------> Convolutional Feature Extractor
\subsection{Convolutional Feature Extractor}\label{sec: feature_extractor}
The first module in the proposed solution aims to encode RGB and depth spatial information into the convolutional feature maps having translation-invariant characteristics. The input of our network consists of co-registered RGB and depth images, from which our model first extracts features from the RGB modality using a CNN. The \emph{convolutional feature extractor} consists of five convolutional blocks of a VGG-16~\cite{21} network, where the blocks have 64, 128, 256, 512, and 512 filter maps respectively. Each block is followed by a max pool layer with a kernel size of 2. The output of the fifth block is considered as the extracted RGB convolutional feature maps. Similarly, the depth image is also passed through another set of convolutional blocks with the same architecture, and the feature maps after every convolution block are concatenated together to obtain depth convolutional feature maps. These two feature maps are feed to the next module. We use the pre-trained VGGFace2~\cite{28} weights and fine-tune them on our datasets. To help tune the convolutional extractors, two more auxiliary branches for identity losses for each of the modalities, depth and RGB, are introduced. These branches include two fully connected (FC) layers, each with 1024 nodes, where the number of nodes is equal to the number of classes in each dataset. These two auxiliary branches, shown by dashed lines in Figure~\ref{fig:architecture}, help the network by learning the weights for the convolutional extractors by back-propagating the error through early layers of the network. These two losses are in addition to the main attention loss as described in section~\ref{sec: Classifier}. It is worth noting that these auxiliary branches are only used during the training process, and can be formulated as:  
\begin{equation}
    \begin{aligned}
        L_{RGB} = - \sum_{c=1}^{M} y_{RGB,c}\log(p_{RGB,c})
    \end{aligned}
    \label{eq:rgb_loss}
\end{equation}
\begin{equation}
    \begin{aligned}
        L_{D} = - \sum_{c=1}^{M} y_{Depth,c}\log(p_{Depth,c})
    \end{aligned}
    \label{eq:depth_loss}
\end{equation}
where $y_{RGB,c}$ and $y_{Depth,c}$ represent the output labels for the corresponding input belonging to a certain class $c$ out of $M$ possible classes, and $p_{RGB,c}$ and $p_{Depth,c}$ are the probability scores after passing through the classifier.

%-------> overview Depth-Guided/how it works
\subsection{Proposed Depth-Guided Attention}\label{sec:depth-guided_attention}
\begin{figure*}[!h]
\centering
\includegraphics[width=1\linewidth]{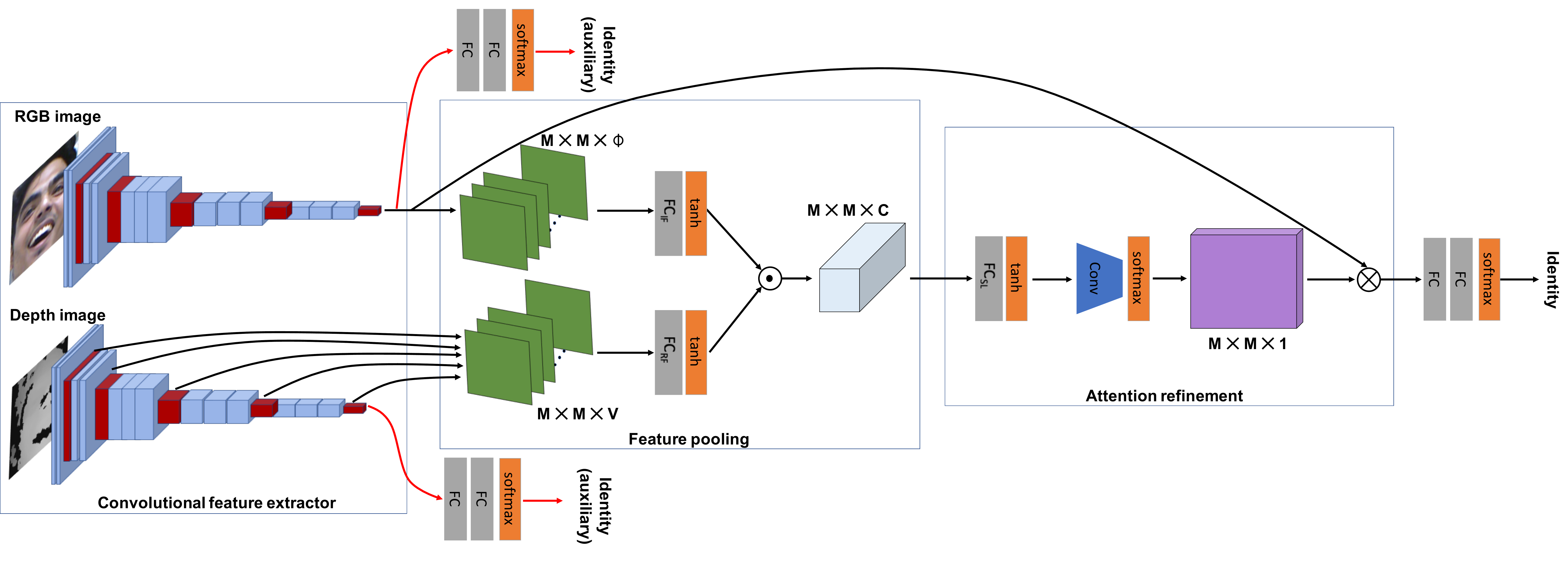}
\caption{Architecture of proposed depth-guided attention network with its two novel modules, feature pooling and attention refinement. Feature pooling finds relations between features in the depth and RGB images, and attention refinement creates a refined attention map for features extracted from the RGB image; the  attention-refined RGB features are finally fed to a classifier.}
\label{fig:architecture}
\end{figure*}

The depth-guided attention mechanism proposed in this paper is composed of the second and third modules, illustrated in Figure~\ref{fig:architecture}, respectively \textit{Feature Pooling} and \textit{Attention Refinement}. The intuition behind the proposed depth-guided attention is to jointly exploit the symmetry between RGB and depth convolutional feature maps to guide the attention mechanism where to look for the most prominent person-specific information within the RGB feature map. 

As discussed in Section~\ref{sec:related_work}-C, some attention mechanisms have used pure RGB features to focus attention~\cite{CBAM,BAM,chen2017sca}, while others~\cite{ICIP,CVPR} have explored the possibility of attention-aware fusion in facial recognition, so as to fuse depth and RGB modalities together. In contrast, our proposed solution (Figure~\ref{fig:attention}) %aims to increase the discriminating power of embeddings. When an RGB image is passed through the CNN, the resulting feature embedding contains spatial information of the image. 
% In this context, our attention mechanism aims to focus on the selected parts of the image, i.e. the depth informative regions of the face such as the lips, chin, forehead, and nose, using depth embeddings to guide the attention on the RGB embeddings. 
multiplies the attention weights derived from the depth-guided attention mechanism by the RGB feature maps extracted from the CNN to obtain the final set of salient features. In the following, we describe Feature Pooling and Attention Refinement modules: 

% Our proposed attention mechanism consists of two modules, labelled \emph{Feature Pooling} and \emph{Attention Refinement}. The Feature Pooling module combines (pools) the features from both RGB and depth modalities to explore the interactions between the two modalities. The subsequent Attention Refinement module further uses the feature space to create an enhanced attention map for the input modality. These modules are described as follows. 

\subsubsection {Feature Pooling Module}\label{sec: Feature_pooling}
The idea behind the feature pooling module is to combine/pool the convolutional feature maps obtained from both RGB and depth modalities to explore the interactions between the two modalities. We investigate two approaches for the Feature Pooling module, the first being bilinear pooling, which has been used in various works~\cite{lin2015bilinear,kim2016hadamard,kim2018bilinear} to record the interactions between features from two modalities. This method provides richer representations of the features compared to linear models as it records all pairwise interactions using outer products between the two modalities' feature maps, i.e., RGB and depth convolutional feature maps. However, as a drawback, bilinear pooling produces a high-dimensional features of quadratic expansion which can tend to be computationally expensive as it uses the outer product between two vectors or tensors, making the rank of the resultant tensor high.
In~\cite{kim2016hadamard}, a less computationally expensive method for bilinear pooling was proposed using the Hadamard product to pool two feature spaces together. The pooled feature spaces created using the first approach can be formulated in the context of our problem as:
\begin{equation}
\begin{aligned}
F_{bp} = \tanh((W_{3})^T(\tanh((W_{1})^T F_{RGB}) \circ \tanh((W_{2})^T F_{D})))
\end{aligned}
\label{eq:bilinear}
\end{equation}
where $W_1$, $W_2$, and $W_3$ are the RGB, depth, and bilinear pooling trainable weights respectively, whereas biases have been ignored for simplicity.
% \(W_{1}\)$\in $ $\mathbb{R}^{m \times n}$, \(W_{2}\) $\in $ $\mathbb{R}^{m \times n}$, and \(W_{3}\)$\in $ $\mathbb{R}^{n \times l}$
% \(W_{1}\) $\in $ $\mathbb{R}^{\phi \times C}$ and \(W_{2}\) $\in $ $\mathbb{R}^{v \times C}$, 
Additionally, $F_{RGB}~\!\!\!\in ~\!\!\!\mathbb{R}^{M \times M \times \phi}$ and  $F_{D}~\!\!\!\in~\!\!\!\mathbb{R}^{M \times M \times V}$ are respectively the RGB and depth convolutional feature maps obtained by the convolution module, where $M$ is the spatial dimension of the feature map in the last convolution block, $\phi$ is the number of feature maps in the last convolution block for the RGB modality and $v$ is the number of concatenated feature maps for each convolution block of depth modality.

The second approach uses the dot product as a similarity measure between the two feature vectors, in the context of our problem the RGB and depth vectors, to create a pooled feature space. This approach has been used in~\cite{xu2016ask} to find the similarity between two vectors. The pooled feature space can be calculated as:
%   Eq. \ref{eq:dot}.
\begin{equation}
\begin{aligned}
F_{dp} = \tanh((W_{1})^T F_{RGB}) \odot \tanh((W_{2})^T F_{D})
\end{aligned}
\label{eq:dot}
\end{equation}
where $\odot$ is the dot product, $W_{1}$ and $W_{2}$ are the trainable weights for RGB and depth respectively, while biases have been ignored for simplicity. The resulting $F_{dp}\!\!\in\!\! \mathbb{R}^{M \times M \times C}$ contains information regarding the correlation between the RGB and depth modalities.

In this paper, we have used the second feature pooling approach (Equation \ref{eq:dot}), as shown in Figure~\ref{fig:architecture}. However, we have also evaluated the performance of our solution adopting the first feature pooling approach (Equation \ref{eq:bilinear}) whose results are presented and compared to the first approach in Section~\ref{sec:ablation_study}.

\begin{figure} [!t]
\begin{subfigure}{0.5\columnwidth}
  \centering
  \includegraphics[width=0.8\columnwidth]{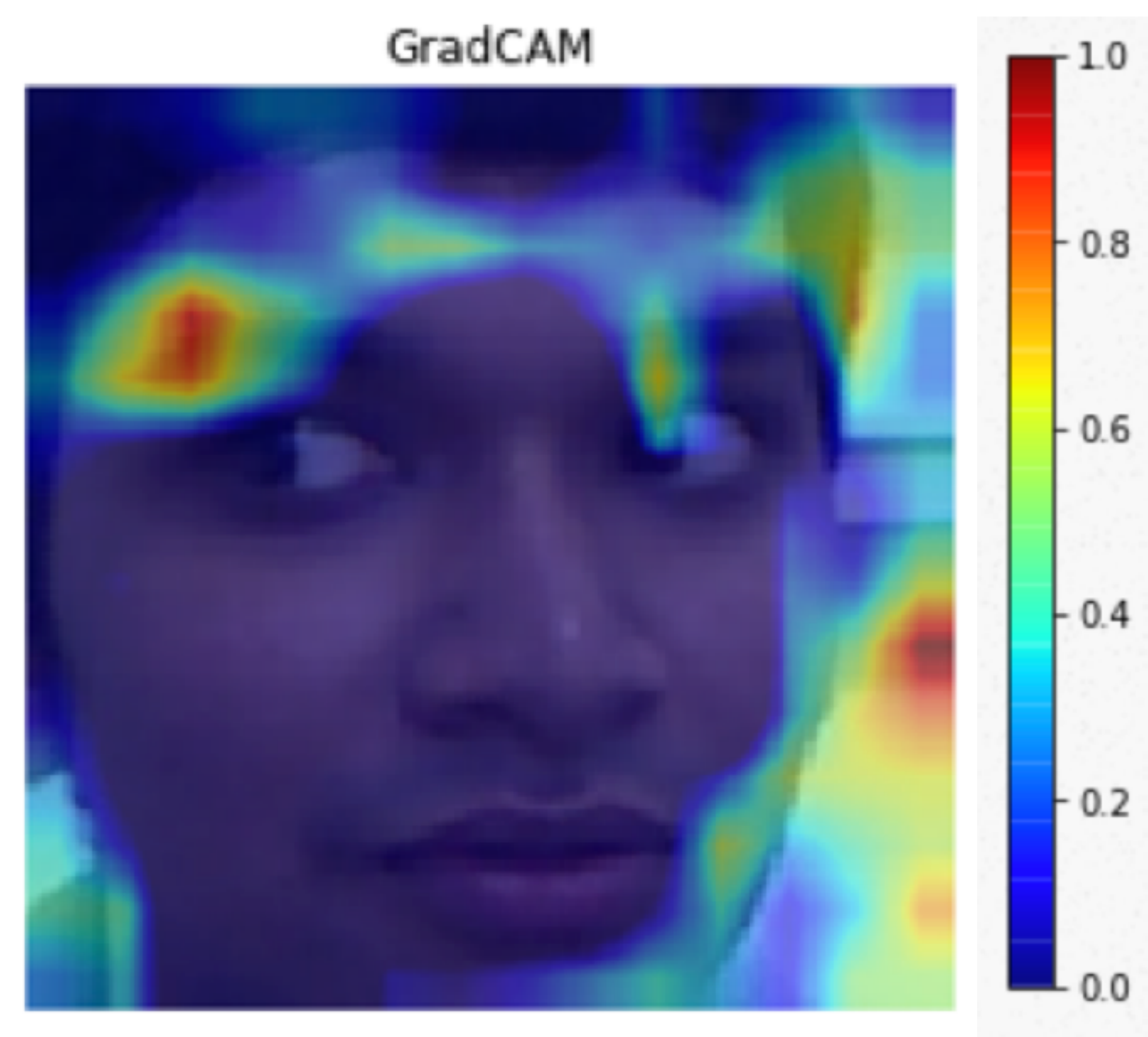}
  \caption{}
  \label{fig:CAM_VGG}
\end{subfigure}%
\begin{subfigure}{0.5\columnwidth}
  \centering
  \includegraphics[width=0.8\columnwidth]{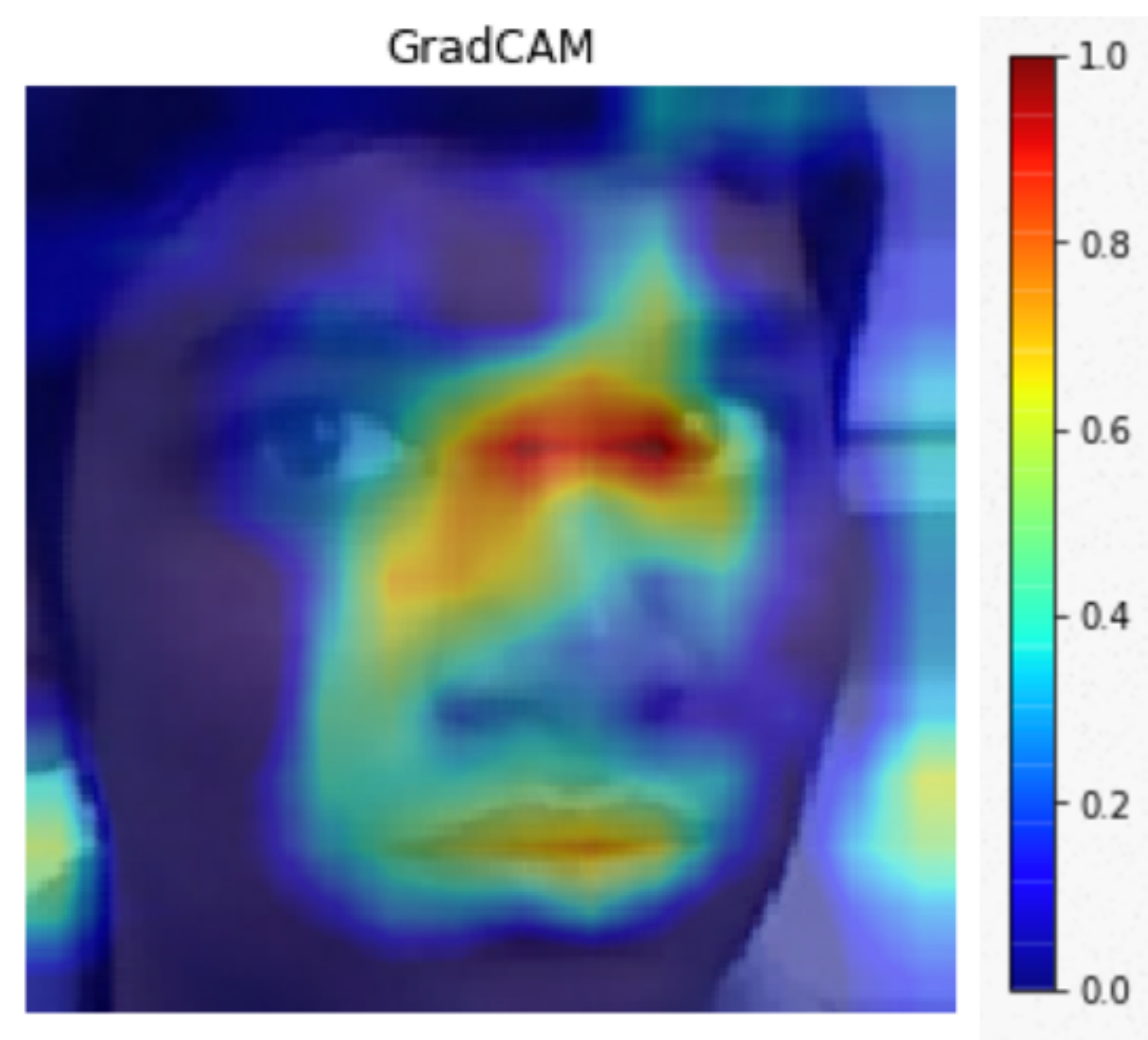}
  \caption{}
  \label{fig:CAM_with_att}
\end{subfigure}
\caption{Class Activation Maps (CAM) for a sample image. (a) shows class activation map obtained by a VGG-16 network with VGG-Face2 weights where the attention is distributed over the image and only few features of the subject's face are highlighted. (b) shows class activation map for the same subject after our proposed attention mechanism is applied; the facial features with higher attention preserve more information about the subject's identity.} 
\label{fig:CAM}
\end{figure}

%-------> Attention refinement
\subsubsection{Attention Refinement Module}\label{sec: attention_refinement}
The tensor resulting from the Feature Pooling module, $F_{dp}$, is used as input to the Attention Refinement module. This module first refines the pooled features by passing its input tensor, $F_{dp}$, through a shared fully connected layer with a tanh activation to add non-linearity to the representations: 
\begin{equation}
\begin{aligned}
F_{sl} = \tanh{((W_{sl})^T F_{dp}),} 
\end{aligned}
\label{eq:conv2}
\end{equation}
where $W_{sl} \in \mathbb{R}^{C \times K}$ is the trainable weight. $F_{sl}$ is then converted to an attention map using a convolution layer with softmax activation:
\begin{equation}
\begin{aligned}
\alpha_{att} = \textit{softmax}(Conv_{1\times1} (F_{sl})
\end{aligned}
\label{eq:conv}
\end{equation}
where$Conv_{1\times1}$ is a convolution layer with kernel size of $1\times1$ and 1 feature map, and $\alpha_{att} \in \mathbb{R}^{M \times M \times 1}$ is the resulting refined attention map. Here $K$ is the number of nodes in the $FC_{sl}$ layer.

A Class Activation Map (CAM) can help visualize the activation of neurons in deep convolution networks. We use GradCAM~\cite{selvaraju2017grad} 
to show the effectiveness of our depth-guided attention in Figure~\ref{fig:CAM}. Figure~\ref{fig:CAM_VGG} represents CAM for a VGG-16 network (pre-trained on VGG-Face2), when only an RGB image is used, while Figure~\ref{fig:CAM_with_att} represents the corresponding CAM for the same subject obtained by the proposed attention-based solution. It can be seen that our solution is able to activate important features of the subject for identification. This contrasts the VGG-16 network output, which distributes the activations over the entire image and not on specific and important facial regions.

\subsection{Classifier}\label{sec: Classifier}
The computed attention weights are multiplied by the RGB embeddings to be finally passed through two fully connected (FC) layers, where the first fully connected layer consists of 2048 nodes and the last fully connected layer has a number of nodes equal to the number of classes. The last FC layer is followed by a softmax activation to obtain the probability of every input image belonging to a certain class. The score for the attention-refined RGB features belonging to a certain class is given by:
\begin{equation}
\begin{aligned}
ID = \textit{softmax}\;((W_{fc2})^T((W_{fc1})^T(F_{RGB} \otimes \alpha_{att}))),
\end{aligned}
\label{eq:classify}
\end{equation}
where $W_{fc1} \in \mathbb{R}^{\phi \times N}$ and $W_{fc2} \in \mathbb{R}^{N \times M}$ are trainable weight parameters for FC layers in the classifier. Here, $\phi$ is the number of feature maps obtained from the last convolution block, $N$ is the number of nodes in the first FC layer of the classifier, and $M$ is the number of classes and nodes of the second FC layer. Subsequently, we can define the loss for the features obtained by our proposed solution as:
\begin{equation}
    \begin{aligned}
        L_{attention} = -\sum_{c=1}^{M} y_{RGB,c}\log(ID)
    \end{aligned}
    \label{eq:att_loss}
\end{equation}
where $y_{RGB,c}$ represent the output vectors of an input belonging to a certain class $c$ out of $M$ possible classes, and $p_{RGB,c}$ and $p_{Depth,c}$ are the probability scores as the output of the classifier.

\subsection{Training Loss}
\label{sec:modality_loss}
Our proposed solution consists of a complementary feature learning approach for improving the fine-tuning process for the RGB and depth convolutional extractors available in the first module. This process has been done using two additional auxiliary branches as identity losses for each of the modalities. The full training loss, including the auxiliary identity losses (Equations \ref{eq:rgb_loss} and \ref{eq:depth_loss}), used to train the entire network, is given by:
\begin{equation}
    \begin{aligned}
        L_{Total} = L_{RGB} + L_{D} + L_{attention}
    \end{aligned}
    \label{eq:total_loss}
\end{equation}

\begin{figure*}[!ht]
  \centering
  \includegraphics[width=0.8\linewidth]{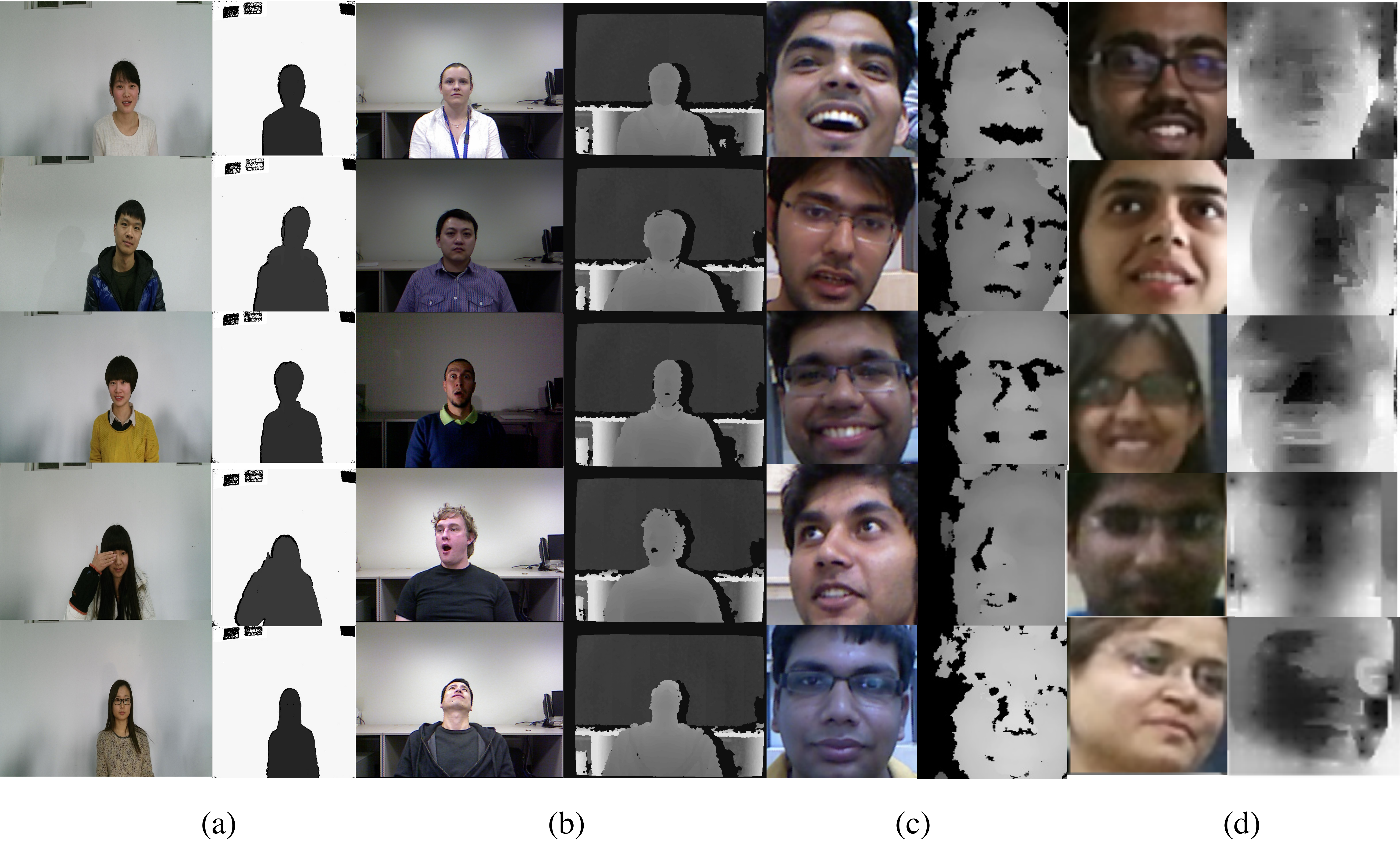}
\caption{Sample images from (a) Lock3DFace~\cite{7}, (b) CurtinFaces~\cite{mian}, (c) IIIT-D~\cite{1,2}, and (d) KaspAROV~\cite{chowdhury2016rgb_newadd}} datasets used in our experiments. 

\label{fig:db}
\end{figure*}

\section{Experiments}
\label{sec:Experiment}
% We evaluate our proposed solution on three publicly available RGB-D datasets, namely Lock3dFace, CurtinFaces, and IIIT-D.
% % , a few samples of which are shown in Figure \ref{fig:db}.
%-------> Datasets - description 
\subsection{Datasets}
\subsubsection{Lock3DFace} The Lock3DFace dataset ~\cite{7} consists of 5671 RGB-D face video clips belonging to 509 individuals with diverse changes in facial expression, pose, occlusion, and time-lapse (Figure~\ref{fig:db}(a)). The dataset has been recorded in two sessions. The neutral images from the first session are considered as the training images and the remaining three variations of the first session form the three test protocols for pose, occlusion, and face expression. The fourth test set consists of all the images from the second session, with all variations. 

\subsubsection{CurtinFaces} CurtinFaces~\cite{mian} is a well-known RGB-D face dataset which contains over 5000 co-registered RGB-D images of 52 subjects, captured with a Microsoft Kinect (Figure~\ref{fig:db}(b)). For each subject, the first 3 images are the frontal, right, and left poses. The remaining 49 images comprise 7 different poses recorded with 7 different expressions, and 35 images in which 5 different illumination variations have been acquired with 7 different expressions. This dataset also contains images with sunglasses and hand occlusions. 

\subsubsection{IIIT-D RGB-D} The IIIT-D RGB-D dataset~\cite{1,2} contains 4605 RGB-D images from 106 subjects captured using a Microsoft Kinect in two sessions (Figure~\ref{fig:db}(c)). Each subject has been captured under normal illumination conditions with variations in pose, expression, and eyeglasses. The dataset already has a pre-defined protocol with a five-fold cross-validation strategy, to which we strictly adhered in our experiments. Each image in the dataset is pre-cropped around the face.

\textcolor{black}{\subsubsection{KaspAROV} The KaspAROV dataset~\cite{chowdhury2016rgb_newadd,chhokra2018unconstrained_newadd} consists of facial videos from 108 subjects recorded by Microsoft Kinect v1 and v2 sensors in unconstrained conditions, as shown in Figure~\ref{fig:db}(d). Each subject appears in two videos that are taken in different acquisition sessions. The dataset includes a total of 432 videos consisting of 117,831 images/frames. 
% The Kinect v2 sensor data in the KaspAROV dataset consists of 62,120 face images of 64$\times$64 resolution. 
The evaluation protocol defined in~\cite{chowdhury2016rgb_newadd} only used the Kinect v2 data due to better registration of its RGB and depth images as compared to the Kinect v1 sensor data (Figure~\ref{fig:db}(d)).}

% \textcolor{black}{
% \subsubsection{VAP RGB-D-T} The VAP RGB-D-T dataset~\cite{nikisins2014rgb_d_t} contains co-registered RGB, depth, and thermal images from 51 subjects as shown in Figure~\ref{fig:db_thermal}. The RGB and depth images were captured using Microsoft Kinect and thermal images were captured at the same time using an AXIS Q1922 camera. The dataset covers 3 variations in pose, illumination, and expression for each subject. The evaluation protocol has been defined with the dataset, splitting the available data into training, validation, and testing sets. }

% \begin{figure}
%     \centering
%     \includegraphics[width=0.5\linewidth]{images/kasparov.pdf}
%     \caption{\textcolor{black}{Sample images from KaspAROV dataset}}
%     \label{fig:db_kasparov}
% \end{figure}

\begin{table}[!h]
    \centering
    \caption{Summary of the test protocol used in our experiments (N~-~Neutral; PS~-~Pose; OC~-~Occlusion; IL~-~Illumination; FE~-~Facial expression; S1~-~Session 1; S2~-~Session 2). }
    \vspace{2mm}
    \begin{tabular}{l c c }
      \hline
      \textbf{Dataset} & \textbf{Gallery}  & \textbf{Test} \\
      \hline
      \multirow{4}{*}{Lock3DFace}& \multirow{4}{*}{N-S1 (6)} & OC-S1 (59) \\
      &  & FE-S1 (59)\\
      &  & PS-S1 (59)\\
      &  & S2 (236)\\
      \hline
      \multirow{2}{*}{CurtinFaces}& \multirow{2}{*}{PS, FE, IL (18)} & PS $\times$ FE (39)\\
      & & IL $\times$ FE (30) \\
      \hline
      IIIT-D RGB-D & Predefined (4) & Predefined (17)\\
      \hline
        KaspAROV & Predefined (287) & Predefined (287)\\
      \hline
        % \multirow{3}{*}{VAP}& \multirow{3}{*}{Mixed (150)} & Rot (50) \\
    %   &  & Exp. (50)\\
    %   &  & Ill. (50)\\
    %   \hline
    \end{tabular}
    % \footnote[1] 
    % \begin{tiny} 
    \\
    \vspace{2mm}
    \small
    %* N~-~Neutral; PS~-~Pose; OC~-~Occlusion; IL~-~Illumination; FE~-~Facial expression; S1~-~Session 1; S2~-~Session 2
    % \end{tiny}
    \label{tab:protocol}
\end{table}

% -------> Datasets - protocol 
\subsection{Test Protocols}\label{sec:Protocol}
We followed the pre-defined protocols for testing as described by the respective authors for the four datasets used. For Lock3Dface, neutral images from session 1 contain 60 frames for each subject, from which 6 equally spaced frames are selected for training as was done in~\cite{CVPR}. The remaining images are divided into four test sets, %the first of which contains occluded images, the second containing facial expression variations, the third containing pose variations, and the fourth containing all of the images from the second session.
containing occluded faces, facial images with different expressions, facial images with different poses and all of the images from the second session, respectively. For CurtinFaces, the training set consists of 18 images per subject, containing only one variation of pose, illumination, or expression. The rest of 69 images are divided into two sets consisting of pose-expression variations and illumination-expression variations as described in~\cite{mian}. The IIIT-D dataset is partitioned into pre-defined testing and training images, which we adhere to. KaspAROV dataset consists of a total of 62,120 images, which are divided equally into test and training sets as mentioned in~\cite{chowdhury2016rgb_newadd}. This amounts to 287 training and testing images per subject with unconstrained pose, expression, and illumination. A summary of the test protocols used in this paper are presented in Table \ref{tab:protocol}.

%------------> details for implementing
\subsection{Implementation Details}\label{sec:Implementation}
\textcolor{black}{\subsubsection{Preprocessing}
Before feeding the images to the network, both RGB and depth images are cropped using the dlib CNN~\cite{dlib} face-extractor network. For unprocessed depth images, we determine two depth values that respectively represent the near and far clipping planes of the scene and filter out the scene content that is either too near or too far from the camera, keeping only depth values which represent the face depth data as suggested in \cite{ICIP}. Following this process, we then normalize the remaining content to fall within the values of 0 to 255, thereby making full use of the full dynamic range of the face depth data.}

\subsubsection{Network parameters}
The optimal parameter values to achieve the best recognition performance have been empirically obtained and are summarized in Table~\ref{tab: parameter}. The CNN component of the network follows a VGG architecture with 5 convolution blocks as described in Section~\ref{sec: feature_extractor}, which are initialized with the weights of the model pre-trained on the VGGFace2~\cite{28} dataset containing over 3.3 million face images from more than 9000 distinct identities. This makes the features generated by the model very general and easily adaptable to the new datasets.%, with some fine tuning.

Our proposed attention mechanism contains 3 fully connected layers where there are $C$ nodes for $FC_{RGB}$ and $FC_{D}$ available in the feature pooling module (i.e. the respective fully connected layers for RGB features (Image feature) and depth features (Reference feature)) as shown in Figure~\ref{fig:architecture}. The third fully connected layer in the attention refinement module has $K$ nodes. The values of $C$ and $K$ were determined empirically as it will be mentioned in section~\ref{sec:hyperparam}. The convolution layer in the attention refinement module has a kernel of $1\times 1$ and 1 feature map with softmax activation. Finally, the classifier has 2 final fully connected layers with 1024 units and a fully connected layer with the number of classes as per the dataset.

\textcolor{black}{We use the Adam optimizer with a learning rate of $10^{-5}$ and decrease it by 10\% with every epoch. Our solution is implemented using TensorFlow \cite{tensorflow} with Keras \cite{keras}, and is trained using an Nvidia GTX 1070 GPU.}

\begin{table}[!t]%[!htbp]
\centering
\caption{The optimal values obtained for the proposed depth-guided FR solution.}
\setlength
\tabcolsep{4pt}
\begin{tabular}{ l| l| l}
\hline
\textbf{Module} & \textbf{Parameter} & \textbf{Setting} \\
\hline
\hline
    Convolutional Feature  & Architecture   &  VGG-16  \\
    Extractor  & Pre-trained weights & VGG-Face2 \\
      & Convolution feature size($F_{RGB}$) & $7\times7\times512$ \\
      & Convolution feature size($F_{D}$) & $7\times7\times1472$ \\

    \hline

    Feature Pooling &  Input image feature size &  $7\times7\times512$  \\
      & Input reference feature size & $7\times7\times1472$ \\
      & Output pooled feature size  & $7\times7\times64$ \\
      & \(FC_{RGB}\)($C$) size & 64 \\
      & \(FC_{RGB}\) activation & tanh \\
      & \(FC_{D}\)($C$) size & 64 \\
      & \(FC_{D}\) activation & tanh \\
    \hline

    Attention Refinement &  Input feature size  & $7\times7\times64$  \\
      & Output attention map  & $7\times7\times1$ \\
      & \(FC_{sl}\)($K$) size & 256 \\
      & Convolution kernel size & 1 \\
      & Convolution feature maps & 1 \\
      & Convolution activation & softmax \\
    \hline
    
    Classifier &  Classifier layers  &  2 FC  \\
      & Number of Layer 1 nodes & 1024 \\
      & Number of Layer 2 nodes & No. of classes \\
      & Layer 2 activation & softmax \\
    \hline

     Full Network  & Batch size & 30 \\
      & Loss function & Cross entropy \\
      & Optimizer & Adam \\
      & Learning rate & 0.00001 \\
      & Learning rate decay & 0.9 \\
      & Metric & Accuracy \\
    \hline
\end{tabular}
\label{tab: parameter}
\end{table}

% More number of neurons does not necessarily mean better performance as shown in Table~\ref{tab:ratio}, which presents the average accuracy on the Lock3dFace dataset. It also decrease the total number of parameter required to train. Figure~\ref{fig:curve_c_k} shows the effect of the ratio of the two layers on the training process. The value of C=64 and K=256 is chosen as it provides the most stable training which highest accuracy. } 

% \begin{table}[!t]
% \setlength{\tabcolsep}{6pt} % Default value: 6pt
% \renewcommand{\arraystretch}{1} % Default value: 1
% \centering
% \caption{Affect of C and K value on Lock3dFace}
% \vspace{2mm}
% \begin{tabular}{ l l c }
%   \hline
%   \textbf{C Value}& \textbf{K Value} & \textbf{Accuracy} \\
%   \hline
%       256 & 512& 84.8\%
%   \\
%       128 & 512 & 85.4\%
%   \\
%     64 & 512& 86.2\%
%   \\
%   64 & 256& 87.3\%
%   \\
%   64 & 128 & 82.3\%
%     \\
%     \hline
% \end{tabular}
% \label{tab:ratio}
% \end{table}

\begin{table*}[!ht]
\setlength{\tabcolsep}{5pt} 
\renewcommand{\arraystretch}{1} 
\centering
\caption{Performance comparison on the Lock3DFace dataset.}
\vspace{2mm}
\begin{tabular}{l l l l l l  c c c c c}
  \hline
\multicolumn{6}{l}{\textbf{}} & \multicolumn{5}{c}{\textbf{Accuracy}}  \\
 \cline{7-11}
  \textbf{Ref.}& \textbf{Year}& \textbf{Authors}& \textbf{Feat. Extractor} & \textbf{Classifier}& \textbf{Input} & \textbf{Pose} & \textbf{Expression} & \textbf{Occlusion} & \textbf{Time} & \textbf{Average} \\

  \hline

  \cite{25}& 2016 & He \etal\ &ResNet-50 & FC/Softmax& RGB & 58.4\% & 96.3\%& 74.7\% & 75.5\% & 76.2\% 
  \\
\cite{hu2018squeeze}& 2017 & Hu \etal\ &SE-ResNet-50 &FC/Softmax&RGB & 60.7\% & 98.2\%& 77.9\% & 78.3\% & 78.7\% 
  \\
   \textcolor{black}{\cite{cui2018improving_newadd}}& \textcolor{black}{2018} &\textcolor{black}{Cui \etal}&\textcolor{black}{Inception-v2 (fusion)} &\textcolor{black}{FC/Softmax}&\textcolor{black}{RGB + Depth  }&\textcolor{black}{ 54.6\% }&\textcolor{black}{97.3\%}&\textcolor{black}{ 69.6\%} &\textcolor{black}{ 66.1\% }&\textcolor{black}{ 71.9\% }\\
%   \cite{13}& 2018 & Zhang \etal\ & RGB + Depth & -- & --& --& --& 90.8\%  
%   \\
  \cite{CVPR}& 2019 & Mu \etal\ & CNN-MFCC-SAV & FC/Softmax& Depth + 3D Model & 70.4\% & 98.2\%& 78.1\% & 65.3\% & 84.2\% 
  \\   
   & \textbf{2020} & \textbf{Proposed} &  \textbf{VGG + Depth-guided Att.} &  \textbf{FC/Softmax} &\textbf{RGB + Depth} & \textbf{70.6\%} & \textbf{99.4\%} & \textbf{85.8\%} & \textbf{81.1\%} & \textbf{87.3\%}
  \\
  \hline
\end{tabular}
\label{tab:Lock3dFace}
\end{table*}

%%%%%%%%%%%%%%%%%%%%%%%%%%
\begin{table*}[!ht]
\setlength{\tabcolsep}{6pt} 
\renewcommand{\arraystretch}{1} 
\centering
\caption{Performance comparison on the CurtinFaces dataset.}
\vspace{2mm}
\begin{tabular}{ l l l l l l c c  c}
  \hline

\multicolumn{6}{l}{\textbf{}} & \multicolumn{3}{c}{\textbf{Accuracy}}  \\
 \cline{7-9}
  \textbf{Ref.} & \textbf{Year} & \textbf{Authors} & \textbf{Feat. Extractor} & \textbf{Classifier}& \textbf{Input}  & \textbf{Pose} & \textbf{Illumination} & \textbf{Average} \\

  \hline
    \cite{25}& 2016 & He \etal\ &ResNet-50&FC/Softmax& RGB & 94.4\% & 96.0\%& 95.7\% 
  \\
\cite{hu2018squeeze}& 2017 & Hu \etal\ &SE-ResNet-50&FC/Softmax&  RGB & 97.4\% & 98.2\%& 97.8\% 
  \\
  \cite{mian}& 2013 &  Li \etal\ & Discriminat Color Space Trans. & SRC & RGB + Depth  & 96.4\% & 98.2\%& 97.3\% 
  \\
  \cite{li2016face}& 2016 & Li \etal\ & LBP + Haar + Gabor& SRC &  RGB + Depth & -- & --  & 91.3\% 
  \\
  \cite{29}& 2016 & Hayat \etal\ & Covariance Matrix Rep. & SVM & RGB + Depth  & -- & --& 96.4\%  
  \\
  \cite{ICIP}& 2020 & Uppal \etal\ &VGG + Two-level Att.&FC/Softmax& RGB + Depth  & 97.5\% & 98.9\%& 98.2\% 
  \\

  & \textbf{2020} & \textbf{Proposed} &  \textbf{VGG + Depth-guided Att.} & \textbf{FC/Softmax}& \textbf{RGB + Depth} & \textbf{98.7\%} & \textbf{99.4\%} & \textbf{99.1}\% 
  \\
  \hline
\end{tabular}
\label{tab:CurtinFaces}
\end{table*}

%%%%%%%%%%%%%%%%%%%%%%%%%%
\begin{table*}[!h]
\setlength{\tabcolsep}{6pt} 
\renewcommand{\arraystretch}{1} 
\centering
\caption{Performance comparison on the IIIT-D dataset.}
\vspace{2mm}
\begin{tabular}{ l l l l l l c }
  \hline
  \textbf{Ref.} & \textbf{Year} & \textbf{Authors} & \textbf{Feat. Extractor} & \textbf{Classifier}& \textbf{Input} & \textbf{Accuracy} \\
  \hline
    \cite{25}& 2016 & He \etal\ &ResNet-50&FC/Softmax& RGB & 95.8\%
  \\
\cite{hu2018squeeze}& 2017 & Hu \etal\ &SE-ResNet-50&FC/Softmax& RGB & 96.4\% 
  \\  
  \cite{1}& 2013 & Goswami \etal\ & RISE & Random Forest& RGB + Depth & 91.6\%  \\
  \cite{2}& 2014 & Goswami \etal\ & RISE & Random Forest& RGB + Depth & 95.3\%  \\
  \cite{13}& 2018 & Zhang \etal\ &9 Layers CNN + Inception &FC/Softmax& RGB + Depth  & 98.6\%  \\
  \cite{chowdhury2016rgb_newadd}& 2016 & Chowdhury \etal\ & Autoencoder &FC/Softmax& RGB + Depth & 98.7\% \\
 \textcolor{black}{\cite{cui2018improving_newadd}}& \textcolor{black}{2018} &\textcolor{black}{Cui \etal}&\textcolor{black}{Inception-v2} &\textcolor{black}{FC/Softmax}&\textcolor{black}{RGB + Depth  }&\textcolor{black}{ 96.5\% }\\
  \cite{ICIP}& 2020 & Uppal \etal\ &VGG + Two-level Att.&FC/Softmax& RGB + Depth  & 99.4\% \\
  \textcolor{black}{\cite{lin2020rgb_newadd}}& \textcolor{black}{2020} &\textcolor{black}{Lin \etal}&\textcolor{black}{ CNN} &\textcolor{black}{ Softmax + Assoc. + Discrim. Loss}&\textcolor{black}{RGB + Depth  }&\textcolor{black}{ \textbf{99.7\%} }\\
  
%   \hline
  & \textbf{2020} & \textbf{Proposed} &  \textbf{VGG + Depth-guided Att.} & \textbf{FC/Softmax}& \textbf{RGB + Depth} & \textbf{99.7}\% \\
  \hline
\end{tabular}
\label{tab:IIIT-D}
\end{table*}
%%%%%%%%%%%%%%%%%%%%%%%%%%%%%%

\begin{table*}[!t]

\centering
\caption{\textcolor{black}{Performance comparison on KaspAROV RGB-D dataset.}}
\vspace{2mm}
\begin{tabular}{ l l l l l l c }
  \hline
  \textbf{Ref.} & \textbf{Year} & \textbf{Authors} & \textbf{Feat. Extractor} & \textbf{Classifier}& \textbf{Input} & \textbf{Accuracy} \\
  \hline
    \cite{chowdhury2016rgb_newadd}& 2016 & Chowdhury \etal& AE Reconstructed Features & FC/Softmax & RGB + Depth &66.7\%
  \\
     \cite{21} & 2014 & Simonyan~\etal &  VGG-16& FC/Softmax & RGB  & 94.5\%
  \\
     \cite{21} & 2014 & Simonyan~\etal &  VGG-16 (score-fusion)& FC/Softmax & RGB + Depth & 94.6\%
  \\
    \cite{21} & 2014 & Simonyan~\etal&  VGG-16 (feature-fusion)& FC/Softmax & RGB + Depth & 94.1\%
  \\
    \cite{cui2018improving_newadd}& 2018 & Cui~\etal & Inception-v2 (feature-fusion) & FC/Softmax& RGB + Depth & 94.8\%
  \\
%   \hline
  & \textbf{2020} & \textbf{Proposed} & \textbf{VGG +  Depth-guided Att.} & \textbf{FC/Softmax}& \textbf{RGB + Depth} & \textbf{ 95.3}\%
    \\
    \hline
\end{tabular}
\label{tab:rgb_kasprov}
\end{table*}

\begin{table*}[!t]
\renewcommand{\arraystretch}{1.2} % Default value: 1
\centering
\caption{Comparison of multimodal methods on all four datasets.}
\vspace{2mm}
\begin{tabular}{ l | c c c c c | c c c | c | c}
%   \hline
% \multicolumn{1}{l}{\textbf{}} & \multicolumn{9}{c}{\textbf{Accuracy}} \\
% \cline{2-10}
\hline
\multicolumn{1}{l}{\textbf{}} & \multicolumn{5}{c}{\textbf{Lock3DFace}} & \multicolumn{3}{c}{\textbf{CurtinFaces}} & \multicolumn{1}{c}{\textbf{IIIT-D RGB-D}} & \multicolumn{1}{c}{\textbf{KaspAROV}}
\\
%  \cline{2-10}
\hline
  \textbf{Model}  & \textbf{Pose} & \textbf{Exp.} & \textbf{Occ.} & \textbf{Time} & \textbf{Ave.} & \textbf{Pose} & \textbf{Illum.} & \textbf{Ave.} & \textbf{Ave.}& \textbf{Ave.}
  \\

  \hline
    RGB+D Fusion& 56.2\% & 97.6\%  & 79.7\% & 75.7\% & 81.64\% & 92.6\% & 94.2\%& 93.4\% & 95.4\% & 94.5\%
  \\
    Attention-aware Fusion & 58.8\% & 98.2\%  & 82.6\% & 75.6\% & 82.8\% & 97.5\% & 98.9\%& 98.2\% & 99.3\%& \textbf{95.3}\%
  \\
    Caps-attention & 51.4\% & 96.8\%  & 78.9\% & 78.3\% & 80.8\% & 96.4\% & 97.8\%& 97.1\% & 98.1\%& 94.9\%
\\
    Cross-modal Attention & 55.1\% & 97.2\%  & 82.7\% & 75.4\% & 81.6\% & 97.5\% & 98.7\%& 98.1\%  & 96.1\%& 95.1\%
  \\
%   \hline
    \textbf{Proposed}& \textbf{70.6\%} & \textbf{99.4\%} & \textbf{85.8\%} & \textbf{81.1\%} & \textbf{87.3\%} & \textbf{98.7\%} & \textbf{99.4\%} & \textbf{99.1}\% & \textbf{99.7\%} & \textbf{95.3\%}
  \\
  \hline
\end{tabular}
\label{tab:Comp_all}
\end{table*}

\section{Results}
\label{sec:Results}
%We validate our solution by testing it on the 3 public RGB-D face datasets Lock3DFace, CurtinFaces and IIIT-D RGB-D. We compare our results to the current state-of-the-art for those datasets to show the effectiveness of our approach. 
%----> Lock3d comp
\subsection{Performance}\label{sec:performance}
The results for the Lock3DFace dataset are shown in Table~\ref{tab:Lock3dFace}. Mu~\etal \cite{CVPR} use depth maps and 3D models to identify the subject in the image and reach an average accuracy of 84.2\%. Specifically, in the presence of pose variations, we achieve an identification rate of 70.6\% which is marginally ($+0.2\%$) better than~\cite{CVPR}. When different facial expressions are tested, our solution achieves an accuracy of 99.4\%, thus obtaining a $+1.2\%$ performance gain. Moreover, when occlusions are applied by subjects covering their faces with certain objects like hands and glasses, we achieve an identification rate of 85.8\% as opposed the best existing reported results of 78.1\%. In the recognition over time scenario, we achieve an accuracy of 81.1\%, which is considerably better than the 65.3\% reported in~\cite{CVPR}. The state-of-the-art SE-Net CNN model~\cite{hu2018squeeze} achieved a maximum accuracy for time variation of 78.3\%, which our solution exceeds by $+2.8\%$. As shown in Table~\ref{tab:Lock3dFace}, our solution achieves an average accuracy of 87.3\%, outperforming the state-of-the-art by $+3.1\%$.

%----> CurtinFaces comp
The results for the CurtinFaces dataset are reported in Table~\ref{tab:CurtinFaces} where our proposed solution achieves state-of-the-art recognition rates in all the test scenarios. Compared to \cite{ICIP}, the best performing benchmarking method, our approach achieves higher ($+0.3\%$) accuracy of 98.7\% for pose variations, while for variations in illumination, our solution achieves near perfect results with an identification accuracy of 99.4\%. The overall average accuracy of our solution is 99.1\% for this dataset, $+0.9\%$ higher than the best-performing alternative method.

%----> IIIT comp
Table~\ref{tab:IIIT-D} shows the results for the IIIT-D dataset.
The results of our solution is slightly better ($+0.3\%$) than the results obtained in \cite{ICIP}, which uses 2-step attention to merge the multimodal embeddings prior to classification. Our method also performs $+1.0\%$ better than \cite{chowdhury2016rgb_newadd} which uses depth-rich features acquired from an autoencoder to obtain a classification accuracy of 98.7\%, and $+1.1\%$ better than \cite{7}, which uses complimentary feature learning to achieve 98.6\% accuracy. \textcolor{black}{It also outperforms the results presented in \cite{cui2018improving_newadd} by $+3.5\%$, which utilizes feature fusion with an Inception-v2 CNN for each modality. Finally, our results are comparable with the current state-of-the-art RGB-D FR method~\cite{lin2020rgb_newadd}}.

%----> KaspAROV

\textcolor{black}{
% Real-world application of RGB-D solution depends on the availability of co-registered RGB and depth modality which is not necessarily the case. 
Lock3DFace, CurtinFaces, and IIIT-D RGB-D datasets contain challenging testing conditions with extreme pose, illumination and expression. However, these datasets have been collected in constrained environments. Table~\ref{tab:rgb_kasprov} shows the results for the KaspAROV \cite{chowdhury2016rgb_newadd} dataset which is collected in unconstrained conditions. The experiments have been performed following the protocol described in~\cite{chowdhury2016rgb_newadd}. The results show that our proposed method performs better than other solutions, achieving an accuracy of 95.3\%, revealing the added value of our depth-guided approach when performing FR in unconstrained conditions.
}

\begin{figure*}[!ht]
  \centering
  \includegraphics[width=0.7\linewidth]{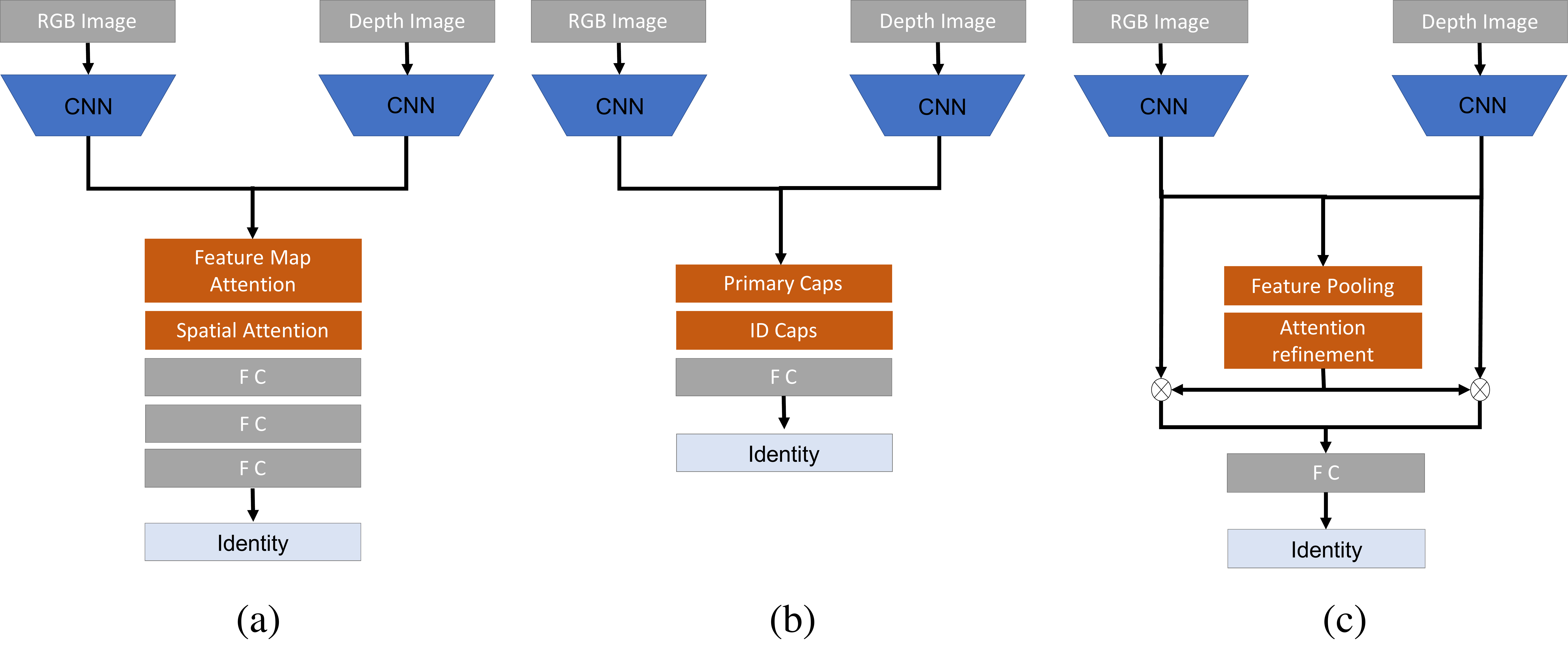}
\caption{Architecture of different multi modal architectures (a) Attention-aware fusion \cite{ICIP}, (b) Capsule attention and (c) Cross-modal attention used in our experiments.}
\label{fig:att_arch}
\end{figure*}

\begin{figure*}[!t]
\begin{subfigure}{.25\textwidth}
  \centering
  \includegraphics[width=\linewidth]{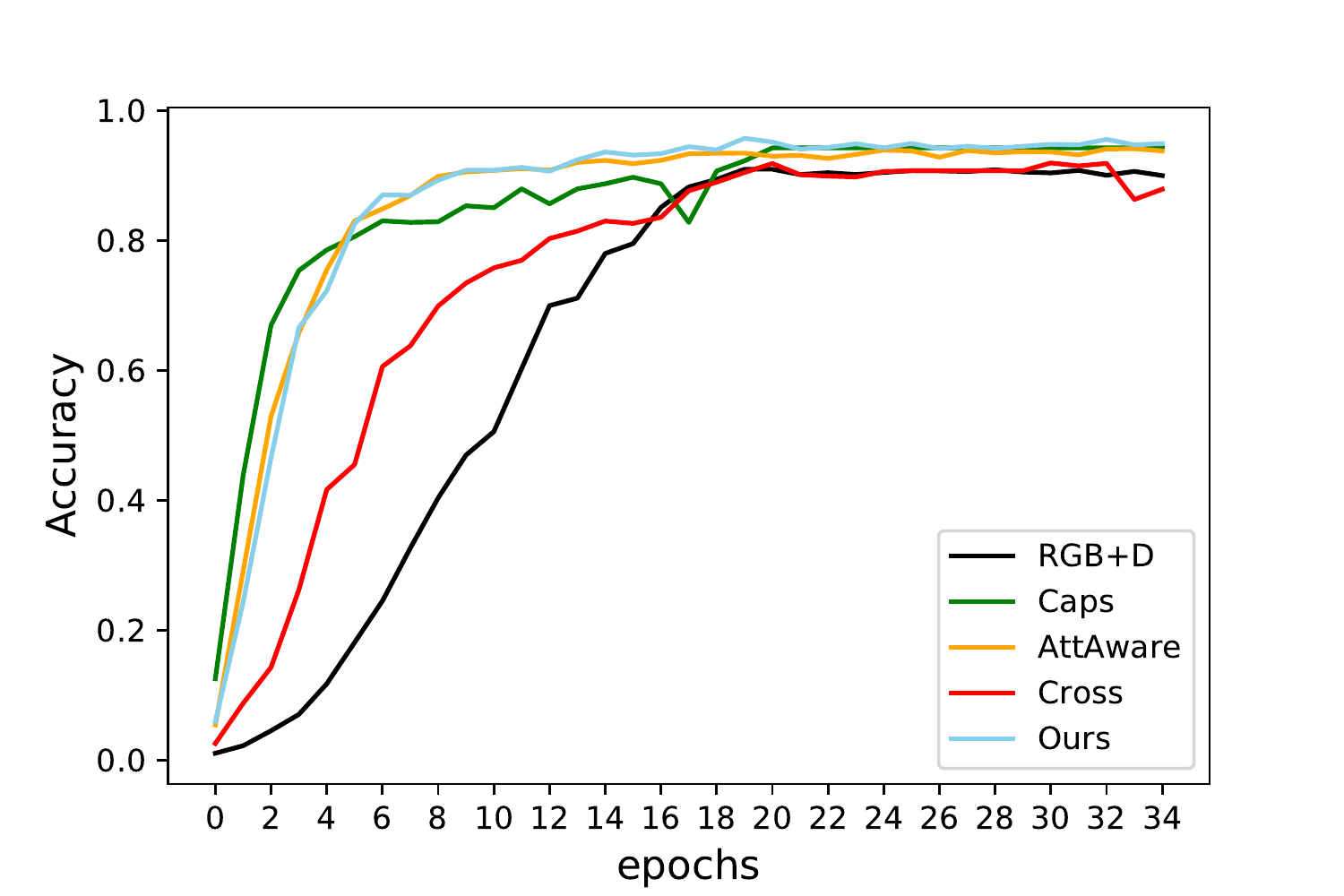}
  \caption{}
  \label{fig:lock3d_cruve}
\end{subfigure}
\begin{subfigure}{.25\textwidth}
  \centering
  \includegraphics[width=\linewidth]{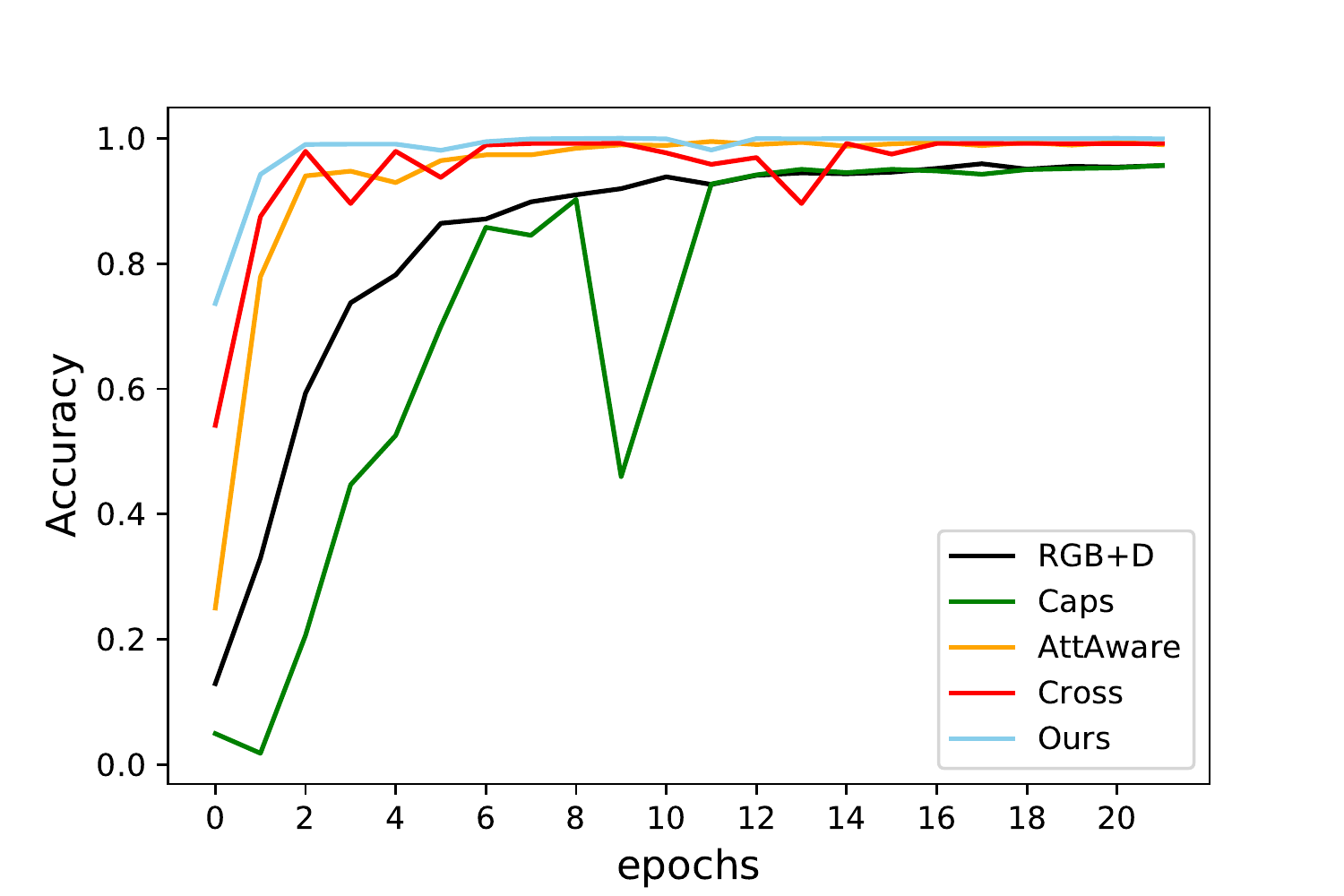}
  \caption{}
  \label{fig:Curtin_curve}
\end{subfigure}%
\begin{subfigure}{.25\textwidth}
  \centering
  \includegraphics[width=\linewidth]{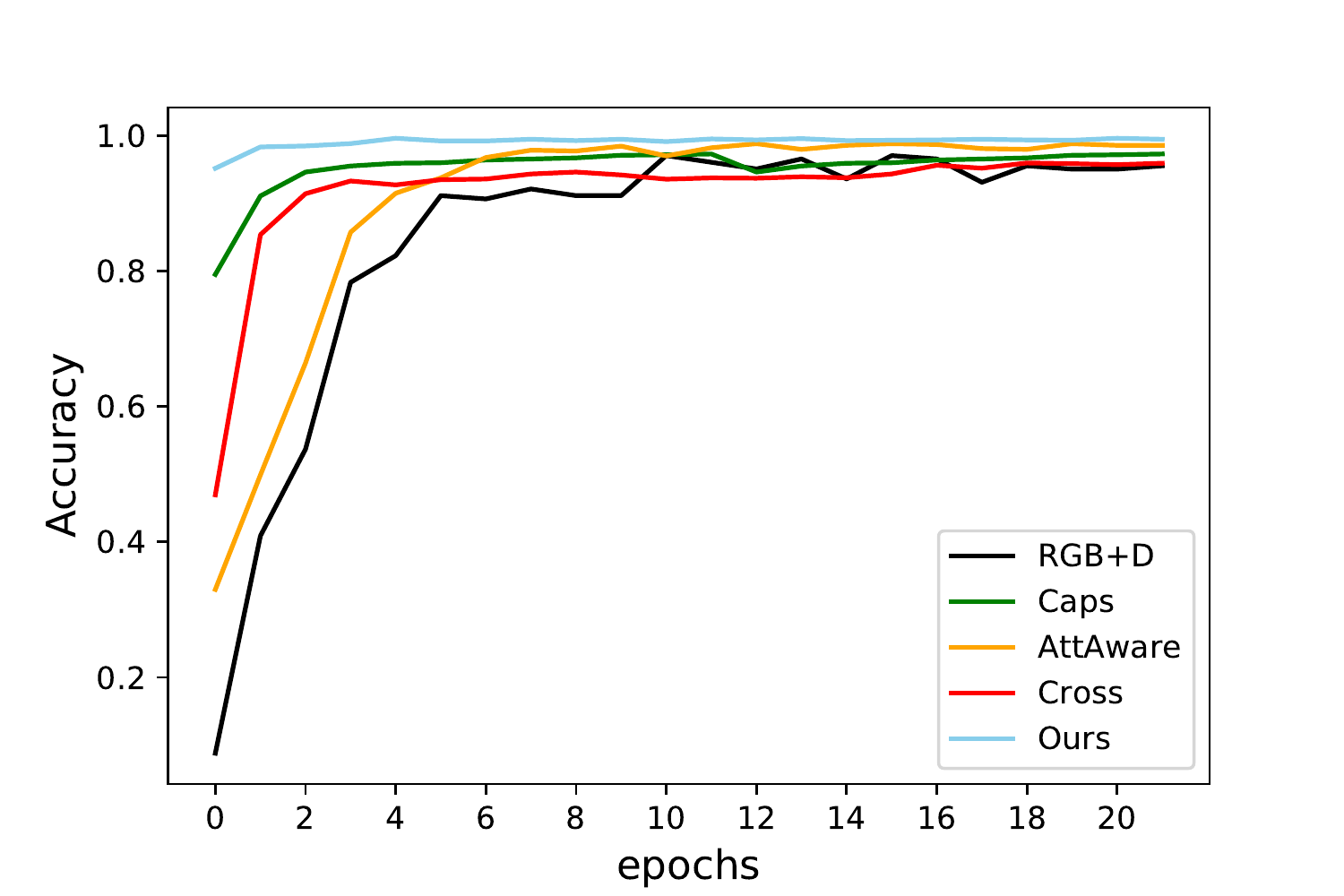}
  \caption{}
  \label{fig:IIIT_curve}
\end{subfigure}%
\begin{subfigure}{.25\textwidth}
  \centering
  \includegraphics[width=\linewidth]{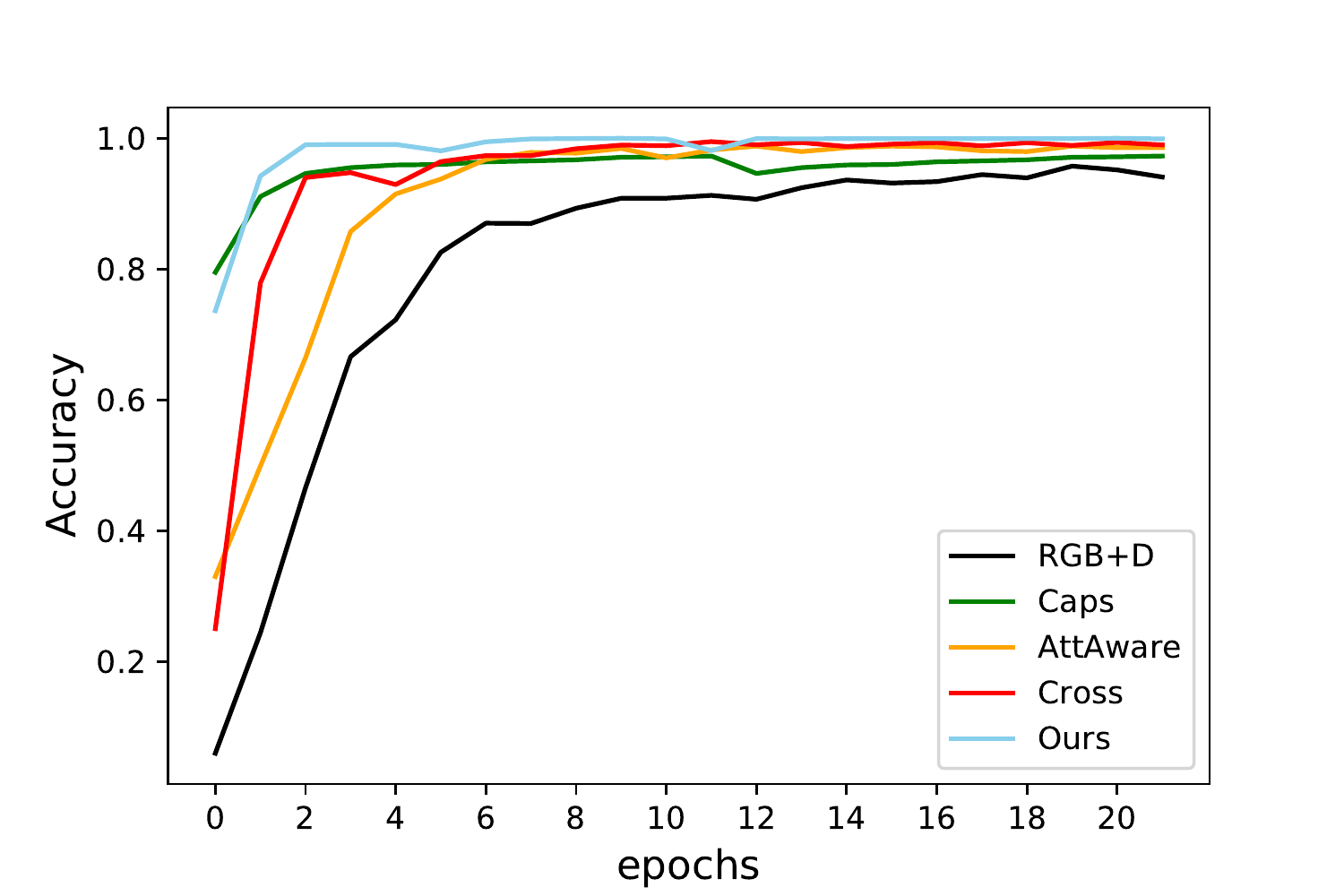}
  \caption{}
  \label{fig:kasparov_curve}
\end{subfigure}%
\caption{Training curves for various multi modal methods on (a) Lock3DFace, (b) CurtinFaces, (c) IIIT-D RGB-D, and (d) KaspAROV datasets.}
\label{fig:curve}
\end{figure*}

%%-- other multimodal approaches

\subsection{Comparison to Other Attention Mechanisms}
We compare the performance of our proposed attention mechanism here with other mechanisms for RGB-D FR, namely attention-aware fusion mechanism~\cite{ICIP}, capsule attention~\cite{sabour2017dynamic}, and cross-modal attention (described below). The results are presented in Table \ref{tab:Comp_all}. 

% Additionally, Figure~\ref{fig:curve} illustrates training curves for various RGB-D attention mechanisms using Lock3DFace, CurtinFaces, and IIIT-D RGB-D datasets.

\subsubsection{Comparison to RGB-D Fusion}
In this fusion strategy, we na\"ively fuse the features extracted from convolution branches by concatenating them. The results show our method considerably outperforms this fusion scheme. %Subsequently, we feed these features to the classifier layer. We compare with naive RGB-D feature fusion as a baseline for all the multimodal approaches.

\subsubsection{Comparison to Attention-Aware Fusion}
Attention-aware fusion architecture~\cite{ICIP} is illustrated in Figure~\ref{fig:att_arch}(a). Depth and RGB features extracted from the convolution block are concatenated and then fed to the attention mechanism, which consists of two steps including feature map attention and spatial attention. The outcome is then fed to the classifier block consisting of 3 fully connected layers. From Table \ref{tab:Comp_all} it can be seen that our attention mechanism achieves better results compared to this benchmark approach. Figure~\ref{fig:curve} shows the training curves for various RGB-D attention mechanisms using the four datasets. Attention-aware fusion converges faster for Lock3DFace but lags in other three datasets.

\subsubsection{Comparison to Capsule Attention}
Following the introduction of capsule networks with dynamic routing by Sabour \etal \cite{sabour2017dynamic}, capsule networks have offered a new avenue for finding relations between convolution features and their part-whole relations using routing-by-agreement. Capsules have also been explored in reference to attention in various areas, ranging from language~\cite{zhang2018attention} to bio-signals~\cite{patrick2019capsule}. We also explore this idea here for pooling the depth and RGB features and finding the interactions between those features using dynamic routing a method similar to ~\cite{zhang2018attention}. We describe the architecture for capsule-based attention in Figure~\ref{fig:att_arch}(b). We use the pooled features from both the depth and RGB streams to form a capsule network and use dynamic routing to find the interaction between those capsules. The results presented in Table \ref{tab:Comp_all} show that our proposed attention mechanism works considerably better than capsule attention for all cases. Nevertheless, capsule attention performs better than na\"ive RGB-D feature fusion on the Lock3DFace and KaspAROV datasets in most testing sets. Figure~\ref{fig:curve} shows that capsule attention converges slowly with respect to other methods across datasets.
% Caps-attention performs on lock3dFace dataset at 80.8\%, and average accuracy for CurtinFaces is at 97.1\%. It also fares at 98.1\% on IIIT-D RGB-D dataset.

\subsubsection{Comparison to Cross-modal Attention}
Several works exploit cross-modal attention \cite{xu2016ask,VQA} to first extract both RGB and depth features and subsequently fuse them to take advantage of the complimentary features in both the modalities. Here we also use a cross-modal attention in which we use the attention map generated in Eq. \ref{eq:conv} and use them to generate attention over both RGB and depth convolution features as shown in Figure~\ref{fig:att_arch}(c). These attention-refined features are then concatenated along the channel axis and fed through the classifier block. To determine the number of layers in the classifier, we conduct a number of experiments. We find that 1 fully connected layer classifier gives the best performance as the corresponding results are shown in Table \ref{tab:Comp_all} as well as Figure \ref{fig:curve}. It can be seen that not only does our proposed attention mechanism perform considerably better that the cross-modal attention, but that it also converges faster. We can also observe that cross-modal attention performs better than other benchmarking attention mechanisms, notably attention-aware and capsule attention mechanisms for the Lock3Dface and CurtinFaces dataset. Nevertheless, our proposed solution always outperforms cross-modal attention for all cases.

\begin{figure*}[!t]
% \hfill
\begin{subfigure}{.33\textwidth}
  \centering
  \includegraphics[width=.8\linewidth]{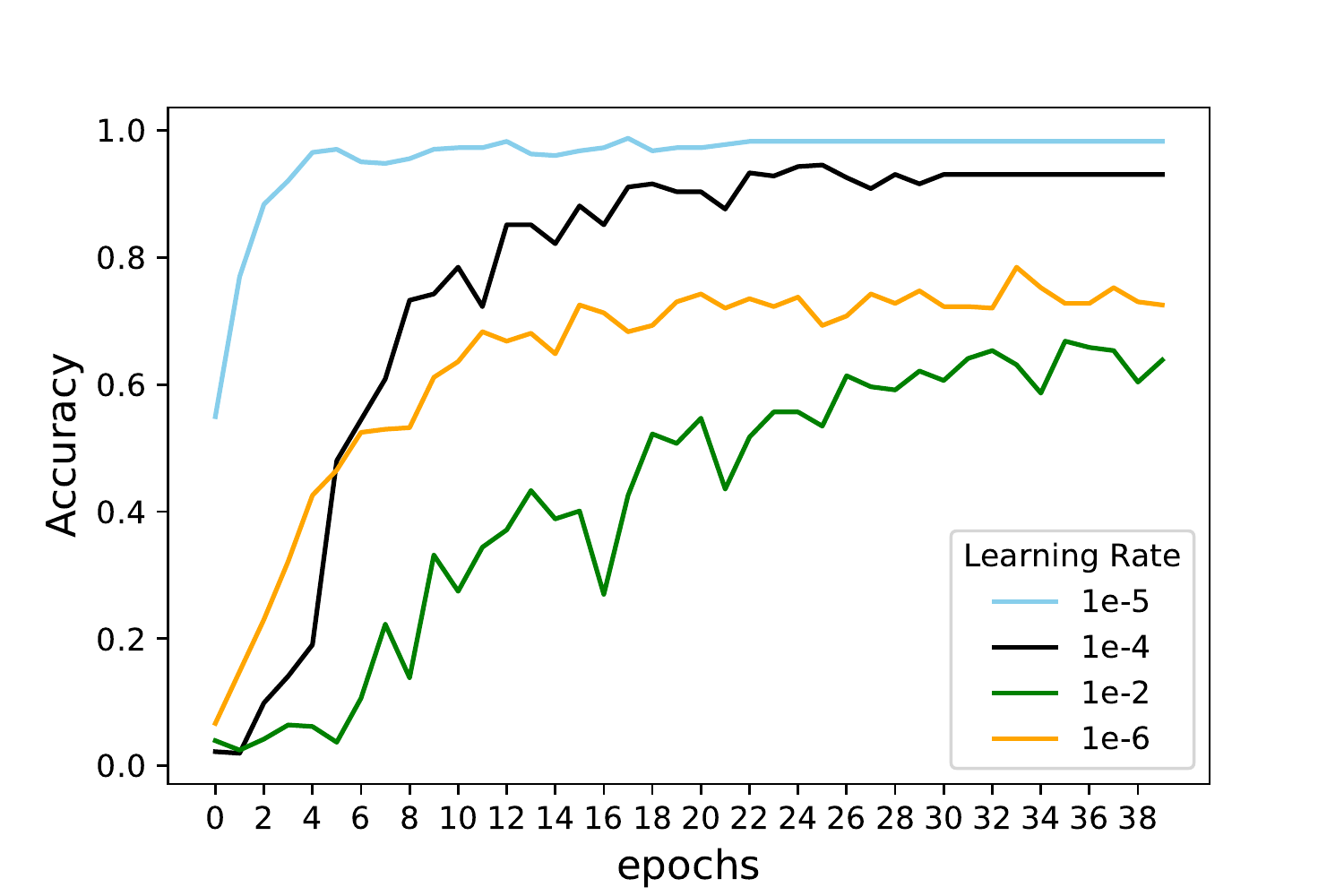}
  \caption{}
  \label{fig:curve_lr}
\end{subfigure}
\begin{subfigure}{.33\textwidth}
  \centering
  \includegraphics[width=.8\linewidth]{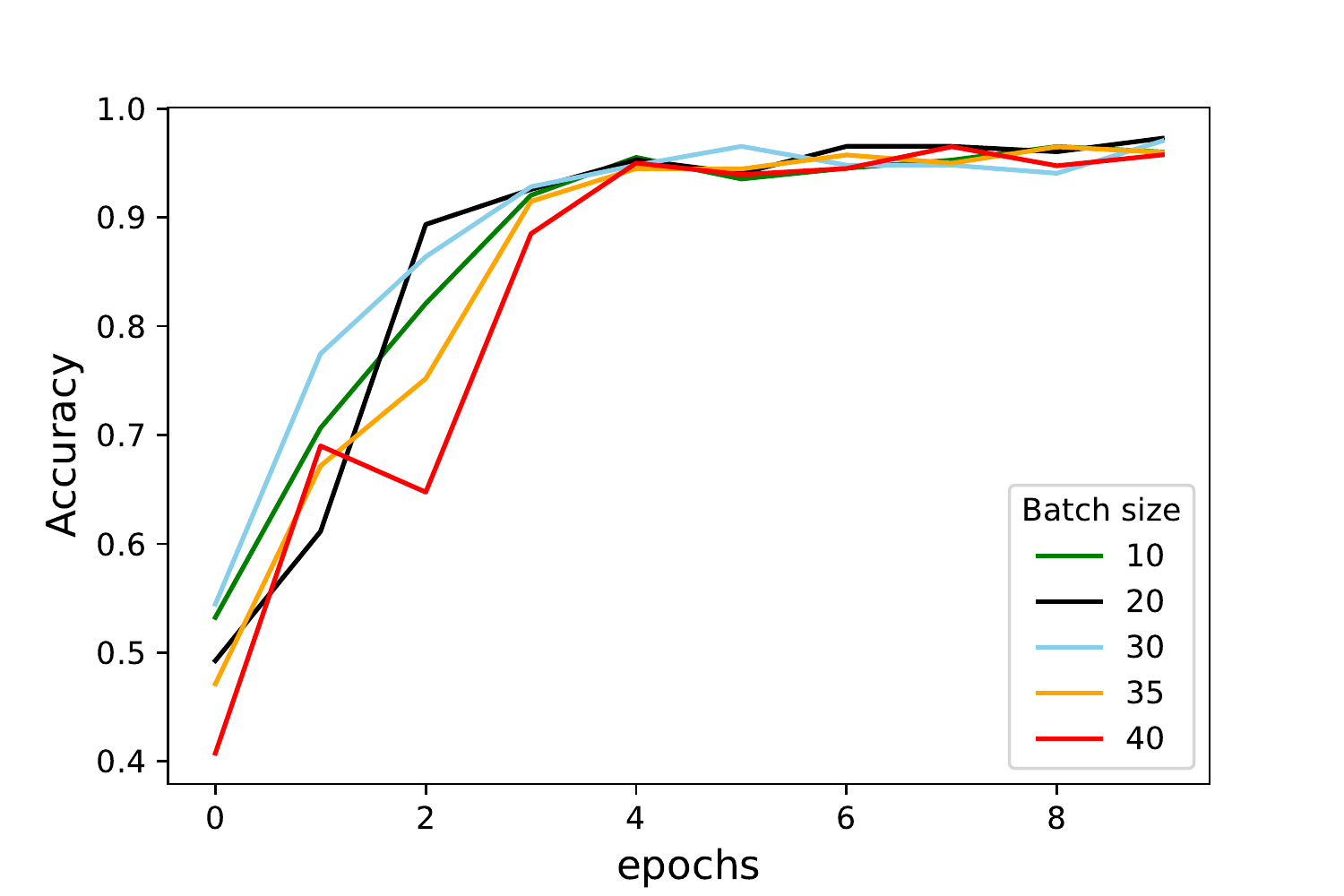}
  \caption{}
  \label{fig:curve_batch}
\end{subfigure}%
\begin{subfigure}{.334\textwidth}
  \centering
  \includegraphics[width=.8\linewidth]{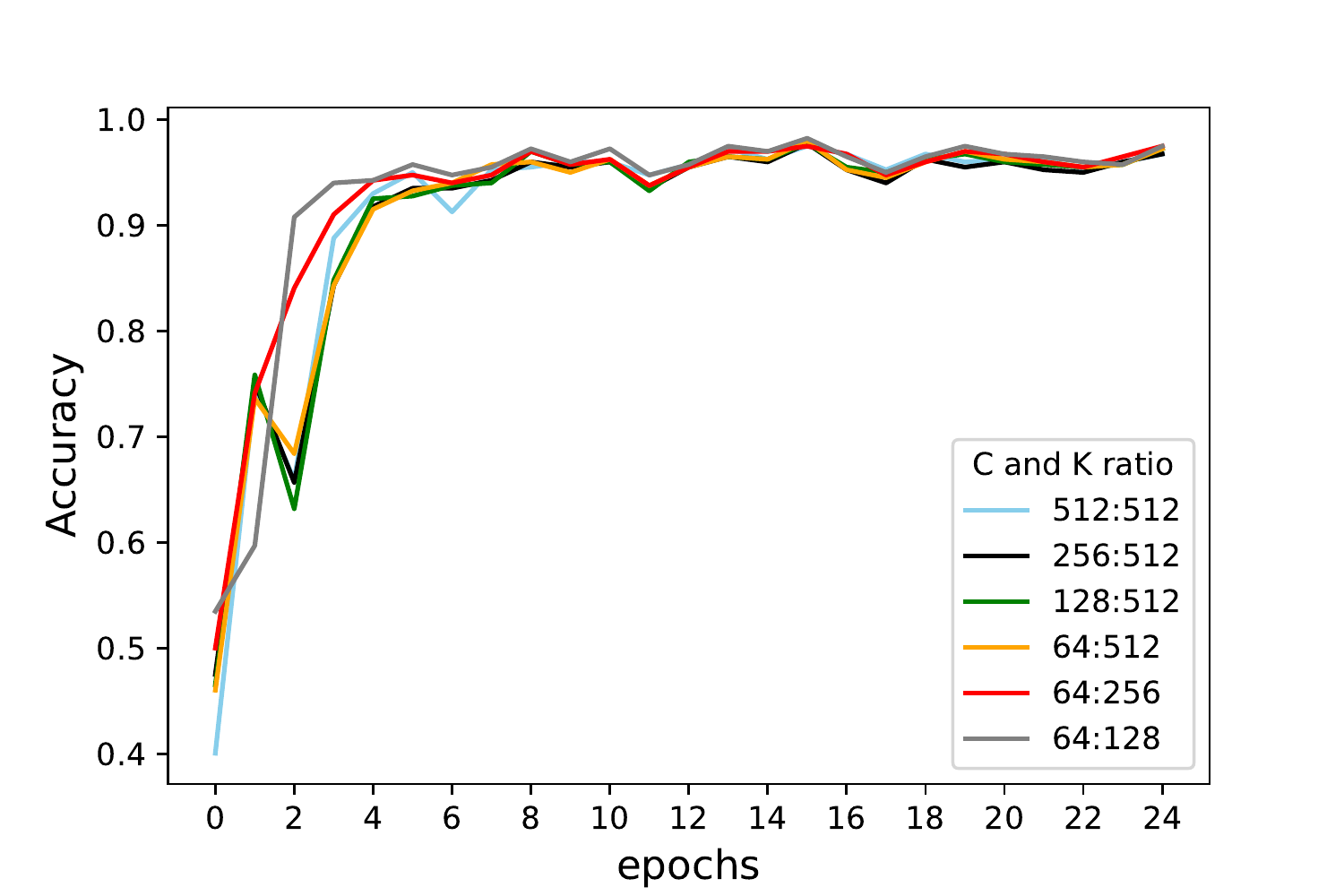}
  \caption{}
  \label{fig:curve_c_k}
%   \hfill
\end{subfigure}%
\caption{\textcolor{black}{Training curves versus various hyperparameters including (a) learning Rate; (b) batch size; and (c) C and K values.}}
\label{fig:curve_hyperparameter}
\end{figure*}

\subsection{Hyperparameter analysis}
\label{sec:hyperparam}
\textcolor{black}{
\textit{Learning rate and batch size:} We conducted extensive experiments on the Lock3DFace dataset to study the sensitivity of our network to learning rate and batch size. The training curves versus learning rate and batch size are respectively shown in Figures~\ref{fig:curve_hyperparameter}(a) and \ref{fig:curve_hyperparameter}(b). Figure~\ref{fig:curve_hyperparameter}(a) shows the sensitivity of training to learning rate, where the learning rate value of $10^{-5}$ enabled the most stable training convergence and hence was chosen as the optimal value. Furthermore, Figure~\ref{fig:curve_hyperparameter}(b) shows that 
% effect of batch size on the training process. Since the effect of batch size does deter the performance much, we can still see 
batch sizes of 20 and 30 provide faster convergence and hence the optimal value of the batch size was chosen as 30.}

\textit{$C$ and $K$ values}: The number of nodes in the fully connected layers in the feature pooling and refinement modules impact the effectiveness of our method, as mentioned in Section~\ref{sec:Implementation}. \textcolor{black}{ 
We can control the complexity of the attention module by varying the number of nodes in the fully connected layers, i.e., $C$ (number of nodes in fully connected layer in feature pooling module) and $K$ (number of nodes in fully connected layer in attention refinement module). Figure~\ref{fig:curve_hyperparameter}(c) shows the effect of these two values on the training process. The values of $C=64$ and $K=256$ are chosen in our method as they provide the most stable and accurate training results.
}

\subsection{\textcolor{black}{Visualization of CAMs}}
\begin{figure}
    \centering
    \includegraphics[width=0.9\linewidth]{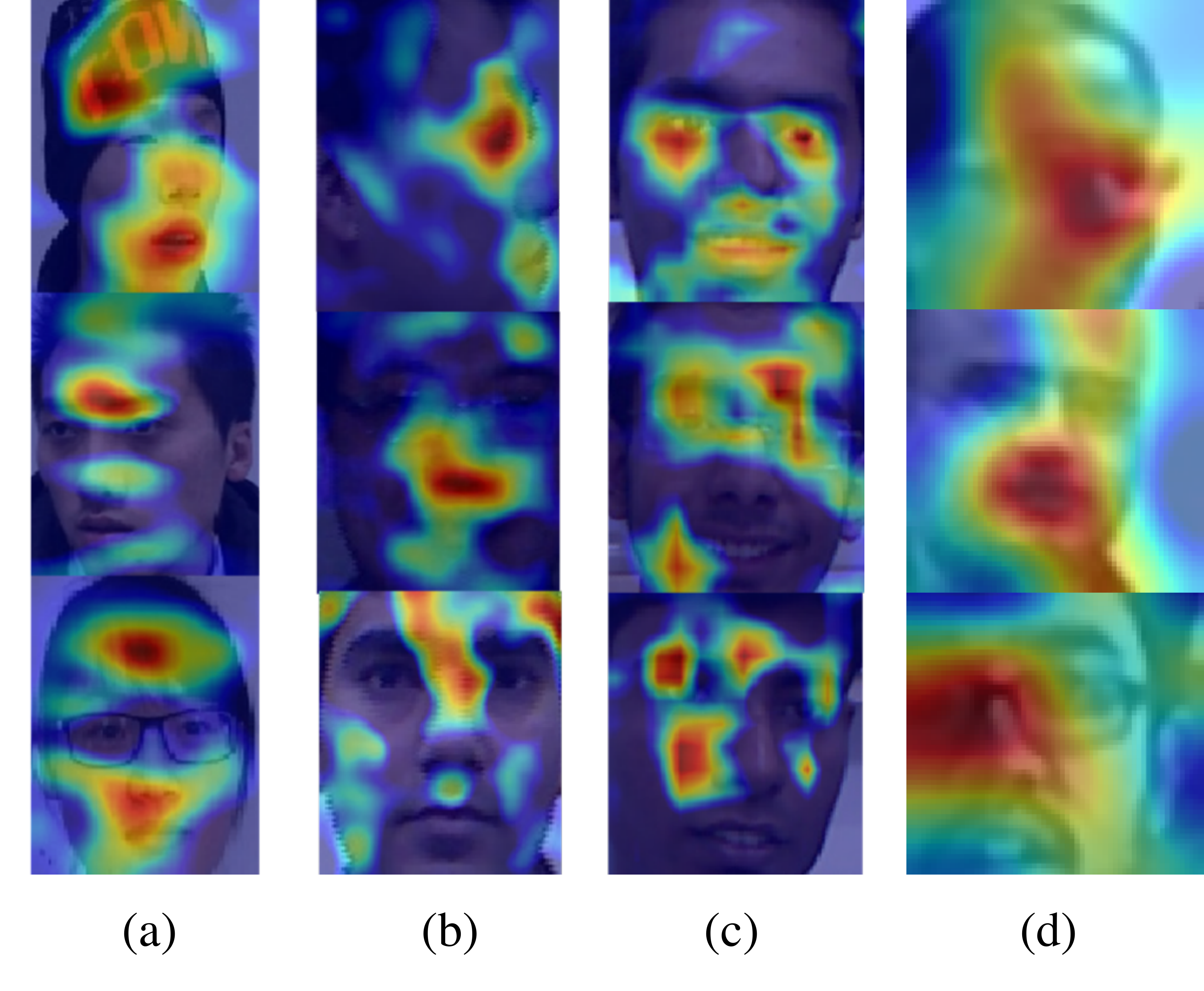}
    \caption{\textcolor{black}{Class activation maps for some samples from four RGB-D face datasets, (a) Lock3DFace~\cite{7}, (b) CurtinFaces~\cite{mian}, (c) IIIT-D~\cite{1,2}, and (d) KaspAROV~\cite{chhokra2018unconstrained_newadd,chowdhury2016rgb_newadd}.}}
    \label{fig:cam_3ds}
\end{figure}

\textcolor{black}{In order to demonstrate the effectiveness of our proposed method and to qualitatively observe the impact of our depth-guided attention, we show several CAMs obtained by our proposed method in Figure~\ref{fig:cam_3ds}. These CAMs are presented for some samples with various pose, occlusion and illumination, from Lock3DFace~\cite{7}, CurtinFaces~\cite{mian}, IIIT-D~\cite{1,2}, and KaspAROV~\cite{chhokra2018unconstrained_newadd,chowdhury2016rgb_newadd} datasets. These results show that the attention maps obtained by our proposed method have selectively focused on the most important regions of the faces.
}

\subsection{Analysis of Failures}
\begin{figure*}[!h]
% \begin{subfigure}{0.25\linewidth}
%     \centering
%     \includegraphics[width=.8\linewidth]{images/lock3d_incorrect.pdf}  
%     \caption{}
%     \label{fig:incorrect_lock3d}
% \end{subfigure}
% \begin{subfigure}{0.25\linewidth}
%     \centering
%     \includegraphics[width=.78\linewidth]{images/curtin_incorrect.pdf}
%     \caption{}
%     \label{fig:correct_curtin}
% \end{subfigure}
% \begin{subfigure}{0.25\linewidth}
%     \centering
%     \includegraphics[width=\linewidth]{images/iiitd_incorrect.pdf}  
%     \caption{}
%     \label{fig:incorrect_iiit}
% \end{subfigure}
% \begin{subfigure}{0.25\linewidth}
%     \centering
%     \includegraphics[width=\linewidth]{images/kasparov_incorrect.pdf}  
%     \caption{}
%     \label{fig:incorrect_kasparov}
% \end{subfigure}
\centering
    \includegraphics[width=0.9\linewidth]{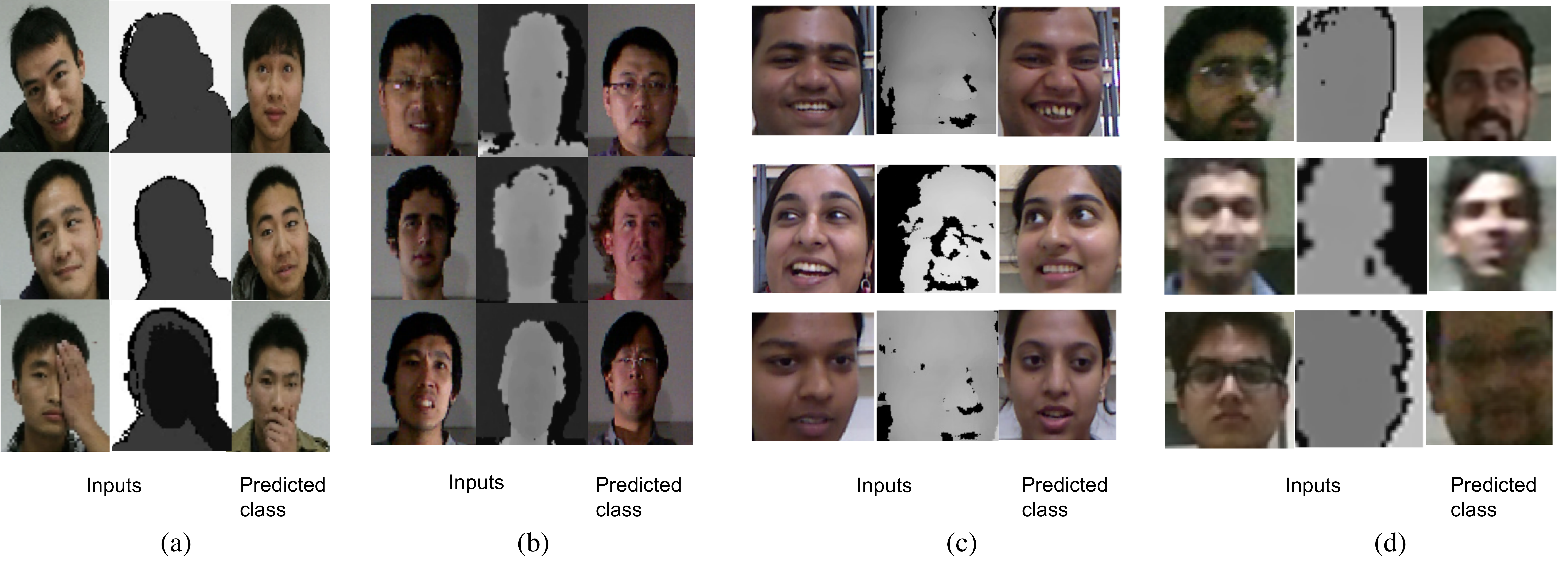} 
\caption{Incorrect predictions on (a) Lock3DFace, (b) CurtinFaces, (c) IIIT-D RGB-D , and (d) KaspAROV dataset for challenging conditions.}
\label{fig:Incorrect_Examples}
\end{figure*}

Figure~\ref{fig:Incorrect_Examples} shows some samples that have been misclassified by our proposed model across the four datasets. For the four datasets, the first two columns on the left show the probe images and the third column on the right corresponds to predicted reference images. It can be observed that the misclassified samples have very similar facial appearances to the probe subjects, and might even be misleading for a human observer. %Also, most of the incorrect predictions were under challenging conditions such as occlusions, dark illumination, or extreme head poses.
% \textcolor{black}{
% \subsection{Analysis on in the wild dataset}
% Real-world application of RGB-D solution depends on the availability of co-registered RGB and depth modality which is not necessarily the case. RGB-D dataset considered in the study contain challenging testing conditions with extreme pose, illumination and expression. Though these datasets were still collected in constrained environments. KaspAROV\cite{chowdhury2016rgb_newadd} dataset collected images in a surveillance type setting which is more unconstrained as compared to other datasets. We perform experiments on KaspAROV dataset with protocol described in ~\cite{chowdhury2016rgb_newadd} and results are shown in Table~\ref{tab:rgb_kasprov}. We can see our proposed method performs as good as other fusion strategies achieving accuracy of 95.3\% . It maybe due to the poor resolution on depth images as seen in Figure~\ref{fig:db_kasparov}
% }

\subsection{Ablation Experiments}\label{sec:ablation_study}
To further study the effectiveness of our proposed solution, we perform extensive ablation experiments with the network on all four datasets. We start with a VGG baseline and observe the model's performance with subsequent added modules.

\subsubsection{VGG-16 baseline}
The VGG-16 network utilized in our solution is first used as the initial baseline. We tested utilizing only this component of the network for identification with \textit{only} RGB images as inputs. We present the results of this experiment in the first row of Table \ref{tab:all_ablation} where average accuracies of 80.9\%, 92.8\%, 94.1\%, and 94.5\% are achieved for the four datasets.

\subsubsection{Pooled feature space}
Next, we explore RGB and depth feature fusion without any attention mechanism. This is done using a combined feature space by concatenating the RGB and depth features from the VGG-16 feature extractor to then feed a 3 layer classifier. We call this \emph{Model A}, which provides a baseline for multimodal FR. As shown in Table \ref{tab:all_ablation}, Model A consistently provides better results than using the RGB modality on its own, with an average increase to 81.6\% ($+0.7\%$) for the Lock3DFace dataset, 93.4\% ($+0.6\%$) for CurtinFaces, and 95.4\% ($+1.3\%$) for the IIIT-D dataset. For the KaspAROV dataset, the concatenation of the RGB and depth features does not increase the accuracy from 94.5\%. 

The proposed attention network was further tested with two additional modified configurations to form the feature pooling module. We define \emph{Model B} with which we test bilinear pooling using the Hadamard product~\cite{kim2016hadamard} for feature pooling, essentially to find the interactions between the two modalities along with modality loss as described in~\ref{sec:modality_loss}. As illustrated in Table \ref{tab:all_ablation}, Model B performs better than Model A in expression and occlusion variations, but lags behind in the second test scenario and under pose variations. The average accuracy for Model B is slightly higher ($+0.7\%$) than that of Model A for the Lock3DFace dataset, with an accuracy of 81.6\%. Model B also performs slightly better than Model A on the KaspAROV dataset. It does however perform significantly better than Model A for the other two datasets, at 98.4\% for CurtinFaces ($+5.0\%$) and 98.5\% for IIIT-D ($+3.1\%$).

\emph{Model C} represents our proposed solution when using the dot product for feature pooling with added non-linearity as explained in Section~\ref{sec:depth-guided_attention}, and modality loss as described in~\ref{sec:modality_loss}. This model does not include the attention refinement module. Model C performs well in all the test settings for the Lock3DFace dataset, as shown in Table~\ref{tab:all_ablation}, obtaining an average identification rate of 85.2\%. Specifically, in the case of occlusions in front of the subject, it performs noticeably better than the three above-mentioned model variations (the baseline, as well as Models A and B). For the CurtinFaces and IIIT-D RGB-D datasets, it performs at 98.5\% and 98.8\% respectively, as shown in Table \ref{tab:all_ablation}, thereby slightly outperforming the previous models, by $+0.3\%$ and $+0.1\%$. For the KaspAROV dataset, Model C performs almost similar to Model B with an accuracy of 94.6\%. 
% It is worth noting that Model B and Model C both also include modality loss as mentioned in ~\ref{sec:modality_loss}.  

\begin{table*}[!h]
\setlength{\tabcolsep}{6pt} % Default value: 6pt
\renewcommand{\arraystretch}{1.2} % Default value: 1
\centering
\caption{Ablation study on all four datasets.}
\vspace{2mm}
\begin{tabular}{ l | c c c c c | c c c | c| c}
 \hline
% \multicolumn{1}{l}{\textbf{}} & \multicolumn{9}{c}{\textbf{Accuracy}} \\
%  \cline{2-10}
\multicolumn{1}{l}{\textbf{}} & \multicolumn{5}{c}{\textbf{Lock3DFace}} & \multicolumn{3}{c}{\textbf{CurtinFaces}} & \multicolumn{1}{c}{\textbf{IIIT-D RGB-D}}& \multicolumn{1}{c}{\textbf{KaspAROV}}
\\
 \cline{2-11}
  \textbf{Model}  & \textbf{Pose} & \textbf{Exp.} & \textbf{Occ.} & \textbf{Time} & \textbf{Ave.} & \textbf{Pose} & \textbf{Illum.} & \textbf{Ave.} & \textbf{Ave.}& \textbf{Ave.}
  \\

  \hline
    Baseline & 55.3\% & 96.4\%  & 78.6\% & 74.1\% & 80.9\% & 92.5\% & 93.2\%  & 92.8\% & 94.1\%& 94.5\%
  \\
    Model A & 56.2\% & 97.6\%  & 79.7\% & 75.7\% & 81.6\% & 92.6\% & 94.2\%  & 93.4\% & 95.4\%& 94.5\%
\\
    Model B & 61.3\% & 97.4\%  & 79.1\% & 77.2\% & 82.8\% & 98.6\% & 98.2\%  & 98.4\% & 98.5\%& 94.8\%
  \\
    Model C & 63.3\% & 98.8\%  & 85.1\% & 79.6\% & 85.2\% & 98.6\% & 98.3\%  & 98.5\% & 98.8\%& 94.6\%
  \\
    Model D & 63.4\% & 98.2\%  & 82.3\% & 75.6\% & 82.3\% & 97.5\% & 98.9\%  & 98.2\% & 99.1\%& 94.6\%
  \\
%   \hline
    \textbf{Proposed}& \textbf{70.6\%} & \textbf{99.4\%} & \textbf{85.8\%} & \textbf{81.1\%} & \textbf{87.3\%} & \textbf{98.7\%} & \textbf{99.4\%} & \textbf{99.1}\% & \textbf{99.7}\%& \textbf{95.3}\%
  \\
  \hline
\end{tabular}
\label{tab:all_ablation}
\end{table*}

% \begin{table}[!h]
% \setlength{\tabcolsep}{6pt} % Default value: 6pt
% \renewcommand{\arraystretch}{1} % Default value: 1
% \centering
% \caption{Ablation study on the CurtinFaces dataset.}
% \vspace{2mm}
% \begin{tabular}{ l c c  c}
%   \hline
% %   \multicolumn{1}{l}{\textbf{Method}} &
% \multicolumn{1}{l}{\textbf{}} & \multicolumn{3}{c}{\textbf{Accuracy}}  \\
%  \cline{2-4}
%   \textbf{Method}  & \textbf{Pose} & \textbf{Illumination} & \textbf{Average} \\
% %  & \textbf{Accuracy} & \textbf{Accuracy}\\
%   \hline
%     \textbf{Baseline}& 92.5\% & 93.2\%  & 92.8\%
%   \\
%     \textbf{Model A}& 92.6\% & 94.2\%  & 93.4\%
% \\
%     \textbf{Model B}& 98.6\% & 98.2\%  & 98.4\%
%   \\
%     \textbf{Model C}& 98.6\% & 98.3\%  & 98.5\%
%   \\
%     \textbf{Model D}& 97.5\% & 98.9\%  & 98.2\%
%   \\
% %   \hline
%   \textbf{Proposed} & \textbf{98.7\%} & \textbf{99.4\%} & \textbf{99.1}\% 
%   \\
%   \hline
% \end{tabular}
% \label{tab:CurtinFaces_abl}
% \end{table}

% \begin{table}[!t]
% \setlength{\tabcolsep}{6pt} % Default value: 6pt
% \renewcommand{\arraystretch}{1} % Default value: 1
% \centering
% \caption{Ablation study on the IIIT-D dataset.}
% \vspace{2mm}
% \begin{tabular}{ l  c }
%   \hline
%   \textbf{Method}  & \textbf{Accuracy} \\
%   \hline
%     \textbf{Baseline}& 94.1\%
%   \\
%     \textbf{Model A}& 95.4\%
% \\
%     \textbf{Model B}& 98.5\%
%   \\
%     \textbf{Model C}& 98.8\%
%   \\
%     \textbf{Model D}& 99.1\%
%   \\
% %   \hline
%   \textbf{Proposed} & \textbf{99.7}\% \\
%   \hline
% \end{tabular}
% \label{}
% \end{table}

\subsubsection{Attention refinement without modality loss}
In order to refine the features from the pooled feature matrix, we obtain a refined feature map using a convolution layer of kernel size 1, and one filter map. To see the effect of the attention refinement module without modality loss, we add the attention refinement module to the best performing model from the above-mentioned variations and remove modality loss to create \emph{Model D}. Now, this model consists of the dot product for feature pooling and the attention refinement module.
% , and the improvement in accuracy in the occlusion and time lapse settings can be observed in
The results in Table~\ref{tab:all_ablation} show marginal reduction in accuracy of expression variation for the Lock3DFace dataset. The degradation is more significant for the other Lock3DFace variations, where the average accuracy 
% decline in accuracy in occlusion and time lapse setting by $2.8\%$ and $4.0\%$. The average accuracy 
decreases by $2.9\%$, when compared to Model C. For the CurtinFaces dataset, we can see a drop in accuracy for pose variation by $1.1\%$ and a marginal performance improvement of $0.6\%$ in illumination variation, compared to Model C. Similarly for the IIIT-D RGB-D dataset, we observe that model D marginally increases the results from 98.8\% to 99.1\%. However, for the KaspAROV dataset, removing the modality loss does not affect the performance. The overall results obtained by Model D show that removing the modality loss degrade performance of our solution. 

% We also marginally increase the accuracy in the cases of pose and expression variations. It improves the results by $+3\%$ in expression variation over the baseline model. A significant improvement can be observed in the more challenging conditions of pose, occlusion, and time lapse settings at 70.6\%, 85.8\% and 81.1\% respectively. The average overall improvement is  approximately $+7\%$ over our baseline results in rank-1 identification rate at 87.3\%, proving the effectiveness of using depth-guided attention. For the CurtinFaces and IIIT-D RGB-D datasets, our results are 99.8\% and 99.7\% respectively, which are very close to perfect scores. Adding the attention refinement module increases the accuracy on the KaspAROV dataset to 95.3\% which is $0.8\%$ higher than the baseline results.

\subsubsection{Attention refinement with modality loss}
To study the joint effect of attention refinement and modality loss, which forms our \emph{proposed solution}, the results are presented in the last row of Table~\ref{tab:all_ablation}. We can see that the complete model increases the accuracy for the Lock3DFace dataset, where 
% It improves the results by $+3\%$ in expression variation over the baseline model. 
a significant improvement can be observed in the more challenging conditions of pose, occlusion, and time lapse settings at 70.6\%, 85.8\% and 81.1\% respectively. The average overall improvement is approximately $+5\%$ over the results of Model D and $+2\%$ over the results of Model C, proving the effectiveness of joint exploitation of attention refinement and modality loss. For the CurtinFaces and IIIT-D RGB-D datasets, our results are 99.1\% and 99.7\% respectively.
% , which are very close to perfect scores. 
Finally, for the KaspAROV dataset, the joint exploitation of attention refinement and modality loss modules improves the result to 95.3\%, which is $0.7\%$ higher than the average performance of both Models C and D.

% that represents our architecture when considering both attention refinment and modality loss modules. 
% but without the modality specific losses for the RGB and depth streams added to the training loss.

% In Lock3dFace we can observe our average accuracy for the dataset increases in our proposed model by $+5\%$ to 87.3\%. Similarly in the CurtinFaces and IIIT-D RGB-D datasets, we observe that model D lags behind the proposed model at 98.2\% and 99.1\% respectively. However, for the KaspAROV dataset, removing the modality loss does not affect the performance.

\subsection{Feature Space Exploration}
\begin{figure*}[!t]
\centering
\includegraphics[width=0.8\linewidth]{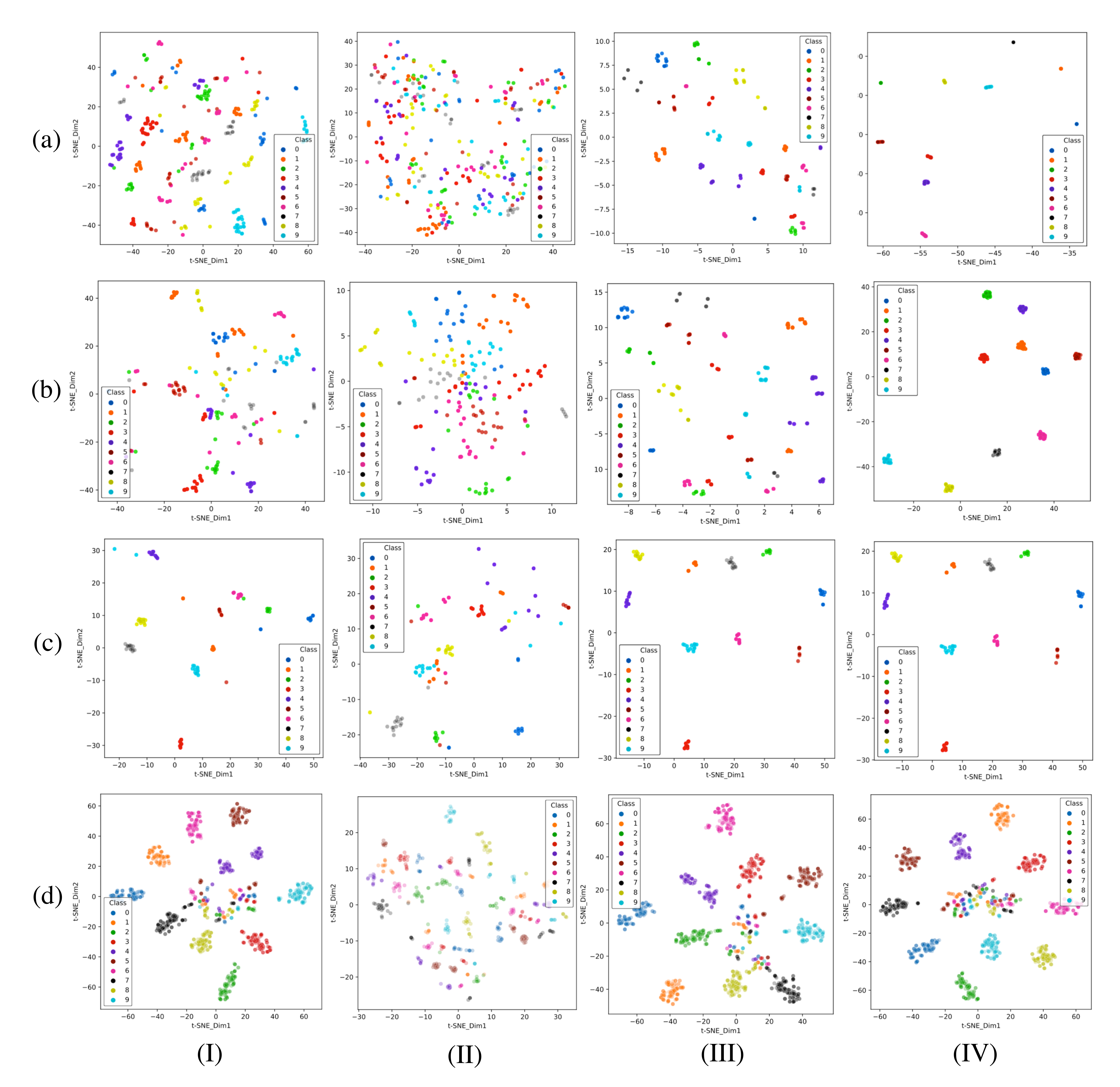} 
\caption{t-SNE visualization of our proposed method and other solutions. Every row in the figure represents a dataset, (a) Lock3DFace, (b) CurtinFaces, (c) IIIT-D RGB-D, and (d) KaspAROV and every column correspond to a solution (I) VGG with only RGB input, (II) VGG with only Depth input, (III) attention-aware fusion, and (IV) proposed depth-guided attention.}
\label{fig:tSne}
\end{figure*}

To explore the impact of our proposed architecture on the feature space, we analyze the space using the t-SNE visualization algorithm~\cite{tSNEvis} by projecting the feature embeddings onto a two dimensional space and observe the discriminative capacity of the learned features. Figure~\ref{fig:tSne}(I) shows embeddings produced by the RGB modality alone for all the four datasets, followed by Figure~\ref{fig:tSne}(II) through (IV) where the depth embedding, the attention-aware fusion embedding~\cite{ICIP}, and the embedding from our proposed solution are visualized respectively. To make the visualisation legible and easy to interpret, we choose 10 subjects from each dataset. We observe that in our solution the subjects form effective clusters. 
% , with the exception of few classes like class 4, 5, 6, and 9, which form distributed clusters that intermingle between themselves in Figure~\ref{fig:tSne}(b).
We could observe similar results across all four datasets in Figure~\ref{fig:tSne}(a), \ref{fig:tSne}(b), \ref{fig:tSne}(c) and \ref{fig:tSne}(d) representing Lock3DFace, CurtinFaces, IIIT-D RGB-D, and KaspAROV datasets, respectively. Through this visual process, it is evident that attention mechanisms help form more distinct clusters than their single modality counter-parts. This effect is more evident in the case of our proposed solution as our solution is able to learn more effective person-specific features and hence improving the classification performance.
% Figure~\ref{fig:tSne}(II) shows the feature embeddings produced by \textit{only} the depth modality. It can be easily observed that in this 2-dimensional feature space, classifying identities only using the depth modality would not be effective due to the lack of clear and separate clusters for all three datasets.

% Figure~\ref{fig:tSne}(III) uses  attention-aware RGB-D fusion~\cite{ICIP} for the same 10 classes. It can be observed that this method, through the effective use of both modalities, is able to form clear clusters for classification. Some of these clusters like classes 4 and 9 have a number of data points which are further away from their respective clusters centres. 

% Figure~\ref{fig:tSne}(IV) represents the embeddings produced after applying our proposed depth-guided attention solution. It is evident that we are able to distinguish between these 10 classes, thereby overcoming the problems posed in the other methods.

% Figures~\ref{fig:tSne}(c) (III) and (IV) represent  attention-aware RGB-D fusion~\cite{ICIP} and our proposed solution respectively. We see that both perform almost equally effectively in clustering the classes, as could also be seen from the identification accuracies obtained.

% \section{Complexity Analysis}
\subsection{Model Complexity and Testing Time Analysis}
\label{sec:Complexity}
\textcolor{black}{The testing time analysis and number of parameters for the proposed method and other benchmarking solutions are provided in Table~\ref{tab:complexity}. The analysis has been done by measuring the execution time on a 64-bit Intel PC with 3.20 GHz Core i7 processor, 16 GB RAM, and Nvidia Geforce GTX 1070 GPU. We used Keras with TensorFlow backend. Table~\ref{tab:complexity} shows the testing time for each RGB image and RGB-D image pairs. We also compare the number of trainable parameters as a measure of complexity for the benchmarking solutions. It is also worth noting that RGB-based solutions used just one CNN stream, while RGB-D based solutions, including ours, used one CNN stream for each modality, thereby increasing the number of trainable parameters. It can be observed that our model has fewer parameters than most of the RGB-D based solutions, except Inception-V2 which has lower accuracy than our model as shown in Table~\ref{tab:Lock3dFace}. Our model also shows faster testing than the other RGB-D solutions, with only 0.007 seconds slower run-time than the VGG-16 network using unimodal RGB images.}

\begin{table}[!t]

\centering
\caption{Average testing times $T$ per image
(in seconds), and the number of parameters 
$N$ (in millions) for
the proposed and benchmarking FR solutions.}
\vspace{2mm}
\begin{tabular}{ l l c c}
  \hline
  \textbf{Solution}&\textbf{Modality} & $T$ (sec.) & $N$ ($\times 10^6$) \\
  \hline
      VGG-16 & RGB & 0.012 & 138
  \\
      Inception v2 & RGB & 0.022 & 24
  \\
      Resnet-50 & RGB & 0.014 & 28
  \\
      SE-Resnet-50 & RGB & 0.017 & 26
  \\ 
    \hline
      VGG-16 (feat. fusion) & RGB+Depth & 0.021 & 256
  \\
     Inception v2 (feat. fusion) & RGB+Depth & 0.032 & 50 
  \\
    Attention-aware Fusion & RGB+Depth & 0.040 & 147  
  \\
    Proposed & RGB+Depth & 0.019 & 132
  \\
    \hline
\end{tabular}
\label{tab:complexity}
\end{table}

\begin{figure}
    \centering
    \includegraphics[width=0.5\linewidth]{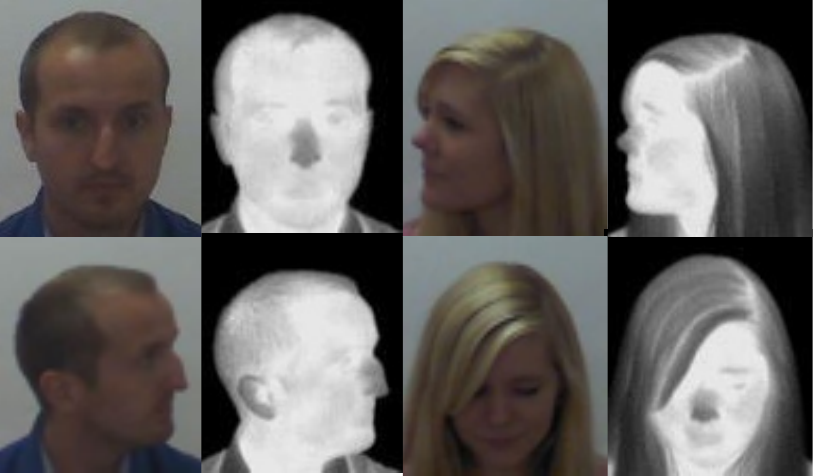}
    \caption{\textcolor{black}{Sample images from VAP RGB-D-T dataset}}
    \label{fig:db_thermal}
\end{figure}

\textcolor{black}{
\subsection{Generalization Study}
We additionally conduct experiments on the VAP RGB-D-T dataset~\cite{nikisins2014rgb_d_t} (Figure~\ref{fig:db_thermal}) to study the generalization of our approach when considering thermal images, instead of depth images, as an auxiliary modality to guide the attention.  
This dataset contains co-registered RGB, depth, and thermal images from 51 subjects as shown in Figure~\ref{fig:db_thermal}. The RGB and depth images have been captured using Microsoft Kinect and thermal images have been captured at the same time using an AXIS Q1922 camera. The dataset covers 3 variations in pose, illumination, and expression for each subject. The evaluation protocol has been defined with the dataset, splitting the available data into training, validation, and testing sets.
Using this dataset, we utilize the thermal images to guide the attention and compare our method with the baseline VGG-16 network applied to the RGB modality as well as score-level and feature-level fusion techniques for RGB and thermal combinations, as presented in Table ~\ref{tab:rgb_thermal}. The results clearly show that our proposed method outperforms all the considered benchmarks, revealing the high generalization ability of our proposed method when adopting other auxiliary modalities (in this case thermal) instead of the depth modality.}

% We use VGG-16 and SE-ResNet-50 with RGB modality 
% as a benchmark for which are trained on 10 images for training and 40 images for validation, we can see SE-ResNet-50 performs better at 90.4\% rank 1- identification accuracy. A naive fusion between SE-ResNet-50 for RGB and Thermal images achieves an accuracy similar to only RGB modality at 90.7\%. We evaluate the combination RGB and Thermal modality with Uppal \etal approach, which uses attention aware fusion between modalities. It achieves and accuracy of 91.3\%. Our proposed method outperforms all these benchmarks with accuracy of 92.1\%.

\begin{table*}[!t]
\centering
\caption{\textcolor{black}{Performance on the VAP RGB-D-T dataset. In this experiment, in order to evaluate the generalization of our proposed method, thermal images are used to guide attention instead of depth.}}
\vspace{2mm}
\begin{tabular}{ l l c c c c}
\hline
\multicolumn{2}{c}{\textbf{ }}&
\multicolumn{4}{c}{\textbf{Accuracy}}
\\
  \cline{3-6}
  \textbf{Method}& \textbf{Input}& \textbf{Expression}& \textbf{Illumination}& \textbf{Rotation} & \textbf{Average} \\
  \hline
      VGG-16 & RGB & 99.3\%& 100\%& 65.4\%& 88.3\% %
  \\
      VGG-16 (score fusion) & RGB + Depth & 99.4\% & 100\%& 65.8\%& 88.4\%
  \\
     VGG-16 (feature fusion) & RGB + Depth & 99.8\%& 100\%& 69.5\%& 89.9\%
  \\
%   SE-ResNet-50 & RGB+Depth & 90.7\%
%   \\
%   Attention Aware Fusion & RGB+Depth & 99.9\%& 100\%& 72.7\%& 90.5\%
%   \\
%   \hline
   \textbf{Proposed} & \textbf{RGB + Depth} & \textbf{99.9\%}& \textbf{100\%}& \textbf{71.6\%}& \textbf{90.3\%}
    \\
    \hline
\end{tabular}
\label{tab:rgb_thermal}
\end{table*}

\section{Conclusion}
\label{sec:Conclusion}
In this paper, we present a depth-guided attention mechanism for RGB-D based face recognition. We extract visual feature embeddings from both depth and RGB modalities and create an attention map for RGB images to increase their classification capability, by guiding attention on specific information-rich areas of the RGB images with the help of the depth modality. Through our evaluations, we validate that our attention mechanism is able to produce more accurate results than the current state-of-the-art on the four public datasets, namely Lock3DFace, CurtinFaces, IIIT-D, and KaspAROV. Further, we test our solution against various multimodal methods like RGB+D fusion, attention-aware fusion, capsule attention, and cross-modal attention, showing that our solution performs better than these architecture variants. We also explore different candidates for the feature pooling module in the ablation studies and find the dot product as a better transformation compared to bilinear pooling using Hadamard product for accentuating person-specific features. Additionally, the experiments with thermal images instead of depth images, show the high generalization ability of our solution when adopting other modalities for guiding the attention mechanism.

In future work, we will explore the performance of our proposed solution using in-the-wild datasets. Moreover, our attention mechanism will also be used with other attribute information to guide the attention of the network towards specific features, depending on the tasks.


%\bibliographystyle{IEEEtran}
\bibliographystyle{ieeetr}
\bibliography{egbib}

\end{document}